\documentclass[a4paper,fleqn]{cas-sc}

\usepackage[authoryear,longnamesfirst]{natbib}

\usepackage{mathtools}
\usepackage{amsfonts}
\usepackage{xcolor}
\usepackage{hyperref}
\usepackage{booktabs}
\usepackage{array,multirow,graphicx}
\usepackage{soul}
\usepackage{arydshln}
\usepackage{float}

\def\tsc#1{\csdef{#1}{\textsc{\lowercase{#1}}\xspace}}
\tsc{WGM}
\tsc{QE}
\tsc{EP}
\tsc{PMS}
\tsc{BEC}
\tsc{DE}

\begin{document}
\let\WriteBookmarks\relax
\def\floatpagepagefraction{1}
\def\textpagefraction{.001}
\shorttitle{BrLP: Brain Latent Progression}
\shortauthors{L. Puglisi et~al.}

\title [mode = title]{Brain Latent Progression: Individual-based Spatiotemporal Disease Progression on 3D Brain MRIs via Latent Diffusion}


\author[1]{Lemuel Puglisi}[type=editor, auid=000, bioid=1, orcid=0009-0003-5661-9873]
\cormark[1]
\ead{lemuel.puglisi@phd.unict.it}
\credit{Conceptualization, Data curation, Formal analysis, Investigation, Methodology, Software, Validation, Visualization, Writing – original draft}

\author[2]{Daniel {C. Alexander}}
\credit{Supervision, Writing – review and editing}

\author[]{for the Alzheimer's Disease Neuroimaging Initiative}
\fnmark[1]

\author[]{the Australian Imaging Biomarkers and Lifestyle flagship study of aging}
\fnmark[2]

\author[3]{Daniele Ravì}
\credit{Conceptualization, Methodology, Project administration, Supervision, Writing – review and editing}


\affiliation[1]{organization={Dept. of Math and Computer Science, University of Catania},
                addressline={Viale Andrea Doria, 6}, 
                city={Catania},
                country={Italy}}

\affiliation[2]{organization={Centre for Medical Image Computing, University College London},
                addressline={90 High Holborn},
                city={London},
                country={UK}}

\affiliation[3]{organization={MIFT Department, University of Messina}, 
                addressline={Viale Ferdinando Stagno d'Alcontres 31},
                city={Messina},
                country={Italy}}

\cortext[cor1]{Corresponding author}

\fntext[fn1]{
Data used in preparation of this article were obtained from the Alzheimer's Disease Neuroimaging Initiative (ADNI) database (adni.loni.usc.edu). As such, the investigators within the ADNI contributed to the design and implementation of ADNI and/or provided data but did not participate in the analysis or writing of this report. A complete listing of ADNI investigators can be found at: \url{http://adni.loni.usc.edu/wp-content/uploads/how\_to\_apply/ADNI\_Acknowledgement\_List.pdf} 
}

\fntext[fn2]{
Data used in the preparation of this article was obtained from the Australian Imaging Biomarkers and Lifestyle flagship study of aging (AIBL) funded by the Commonwealth Scientific and Industrial Research Organisation (CSIRO) which was made available at the ADNI database (\url{www.loni.usc.edu/ADNI}). The AIBL researchers contributed data but did not participate in analysis or writing of this report. AIBL researchers are listed at \url{www.aibl.csiro.au}.
}

\begin{abstract}
The growing availability of longitudinal Magnetic Resonance Imaging (MRI) datasets has facilitated Artificial Intelligence (AI)-driven modeling of disease progression, making it possible to predict future medical scans for individual patients. However, despite significant advancements in AI, current methods continue to face challenges including achieving patient-specific individualization, ensuring spatiotemporal consistency, efficiently utilizing longitudinal data, and managing the substantial memory demands of 3D scans. To address these challenges, we propose Brain Latent Progression (BrLP), a novel spatiotemporal model designed to predict individual-level disease progression in 3D brain MRIs. The key contributions in BrLP are fourfold: (i) it operates in a small latent space, mitigating the computational challenges posed by high-dimensional imaging data; (ii) it explicitly integrates subject metadata to enhance the individualization of predictions; (iii) it incorporates prior knowledge of disease dynamics through an auxiliary model, facilitating the integration of longitudinal data; and (iv) it introduces the Latent Average Stabilization (LAS) algorithm, which (a) enforces spatiotemporal consistency in the predicted progression at inference time and (b) allows us to derive a measure of the uncertainty for the prediction at the global and voxel level. We train and evaluate BrLP on 11,730 T1-weighted (T1w) brain MRIs from 2,805 subjects and validate its generalizability on an external test set comprising 2,257 MRIs from 962 subjects. Our experiments compare BrLP-generated MRI scans with real follow-up MRIs, demonstrating state-of-the-art accuracy compared to existing methods. The code is publicly available at: \url{https://github.com/LemuelPuglisi/BrLP}.
\end{abstract}


\maketitle


\section{Introduction}\label{sec:introduction}

Neurodegenerative diseases represent one of the most pressing challenges in modern healthcare, as these conditions lead to an irreversible decline in brain function and quality of life. With no effective cures available to date, patients and caregivers face prolonged suffering, while healthcare systems struggle with escalating costs and resource demands. Tackling this crisis requires a paradigm shift toward proactive strategies that prioritize early intervention, precision medicine, and comprehensive care. These diseases are notoriously complex, displaying a wide range of neuropathological variations linked to distinct molecular subtypes~\citep{tijms2024cerebrospinal}. Furthermore, they manifest unevenly across brain regions, progressing through diverse mechanisms and at varying speeds, reflecting the intricate nature of their pathophysiology~\citep{young2018uncovering}. Addressing this issue requires the development of advanced tools to deepen our understanding of disease mechanisms, ultimately facilitating the creation of tailored and more effective therapeutic strategies. \newline

Early approaches to disease progression modeling focused on capturing the dynamics of scalar biomarkers~\citep{young2024data,oxtoby2017imaging}. For instance, in~\citep{firth2019longitudinal}, the characteristic atrophy patterns observed in subjects with posterior cortical atrophy are studied by modeling the volumetric changes in specific brain regions. Although scalar biomarkers provide a simplified representation, they have significantly advanced our understanding of various neurodegenerative diseases, such as Alzheimer's Disease (AD)~\citep{vogel2021four} and multiple sclerosis~\citep{eshaghi2021identifying}. However, a significant limitation of these approaches is their inability to capture spatiotemporal characteristics that may more accurately reflect the underlying pathophysiology of a disease. For instance, patients with frontotemporal dementia exhibit shape alterations in the thalamus prior to any detectable volumetric reduction~\citep{cury2019spatiotemporal}. \newline

The growing recognition of spatiotemporal patterns has driven the evolution of traditional disease progression models toward spatiotemporal approaches. Spatiotemporal models~\citep{young2024data} typically leverage high-dimensional data, such as 3D shapes or full medical scans, to represent disease dynamics in a more detailed and holistic manner, enabling the visualization and precise localization of complex structural changes over time. \newline
 
In particular, this paper will focus on spatiotemporal models applied to 3D T1w brain MRIs, with the goal of estimating the structural changes occurring in the brain at the individual level, under both pathological (e.g., neurodegeneration) and non-pathological conditions (i.e., aging). We identify and focus on four primary challenges associated with this task:

\begin{enumerate}
\item \textit{Individualization.} Disease progression is influenced by various individual factors, including demographic and clinical variables. To improve prediction accuracy, models must incorporate and leverage subject-specific metadata.
\item \textit{Longitudinal Data Exploitation.} Longitudinal data offer valuable insights into individual disease trajectories, such as the rate of progression for each patient. When available, models should integrate this data into the inference process.
\item \textit{Spatiotemporal Consistency.} Predictions of disease progression across multiple time points should display a smooth, consistent evolution that aligns with the underlying biological processes.
\item \textit{Memory Demand.} Processing 3D medical images requires significant memory resources, which can limit model applicability in low-resource environments~\citep{blumberg2018deeper}. Enabling such models to run on consumer-grade hardware would support more widespread adoption.
\end{enumerate}

To solve these challenges, we introduce \textbf{Brain Latent Progression} (\textbf{BrLP}), a novel individual-based spatiotemporal model capable of predicting disease progression on 3D brain MRIs at the individual level. BrLP offers several key contributions to address the outlined challenges. First, we propose combining a Latent Diffusion Model (LDM)~\citep{rombach2022high} with a ControlNet~\citep{zhang2023adding} to generate individualized brain MRIs conditioned on available subject data, addressing challenge 1. Second, we integrate prior knowledge of disease progression by employing an auxiliary model that infers volumetric changes in different brain regions, enabling the use of longitudinal data when available and addressing challenge 2. Third, we introduce \textbf{Latent Average Stabilization} (\textbf{LAS}), a technique to improve spatiotemporal consistency in the predicted progression, addressing challenge 3. Fourth, we utilize latent representations of brain MRIs to reduce memory demands for processing 3D scans, addressing challenge 4. Finally, we demonstrate how LAS can be used to derive a measure of uncertainty in the predictions, both at the global and voxel level,  which could serve as a reliability metric in clinical applications. \newline

We train BrLP to learn the structural changes occurring in the brain of subjects with different cognitive statuses: Cognitively Normal (CN), Mild Cognitive Impairment (MCI), and AD. To do so, we use a large dataset of 11,730 T1w MRIs from 2,805 subjects, sourced from three publicly available longitudinal studies on AD. Furthermore, we employ an external longitudinal dataset of 2257 T1w MRIs from 962 subjects to evaluate the generalization capabilities of our method to out-of-sample data. To the best of our knowledge, we are the first to propose a 3D conditional generative model for brain MRI that incorporates prior knowledge of disease progression into the image generation process. \newline

This work extends our MICCAI 2024 conference article~\citep{puglisi2024enhancing} in several ways: (1) we enrich the ablation study by analyzing the hyperparameter for the LAS algorithm; (2) we test BrLP on an external dataset to evaluate its generalization capabilities; (3) we evaluate the impact of the cognitive status as a conditioning variable; (4) we introduce a mechanism to quantify the uncertainty of predictions, both at the global and voxel level, within the BrLP framework and provide statistical analysis to support our findings; and (5) we showcase an example of a potential clinical application of BrLP for patient selection in clinical trials.


\section{Related work}
In this section, we review prior work on spatiotemporal disease progression modeling, which can be broadly categorized into population-based and individual-based approaches. We then focus on the latter, examining how recent studies have leveraged advances in generative AI to produce synthetic scans for individual-level disease progression. Finally, we discuss the limitations of existing approaches and how our proposed method addresses these challenges.

\subsection{Population-based progression modeling}
Population-based progression modeling involves learning an average disease trajectory from longitudinal patient data, often using methods such as mixed-effects models~\citep{schiratti2017bayesian}. Within this framework, individual predictions are obtained by modeling deviations from the average trajectory, allowing for a degree of personalization. A central challenge in such models is mapping individual subjects along a unified disease timeline, as chronological age fails to account for critical inter-patient variations in both age at disease onset and rate of progression. For example, in~\citep{schiratti2015learning}, the authors define a general spatiotemporal model using Riemannian geometry. This method estimates the average disease trajectory as a geodesic on a Riemannian manifold and considers individual trajectories as curves parallel to the average geodesic. The method uses time reparameterization to map individuals to the shared disease timeline. While the Riemannian formulation is highly generalizable, it requires the definition of a suitable Riemannian metric. In a related study~\citep{sauty2022riemannian}, the Riemannian metric is learned from the data, reducing the inductive bias of the model at the cost of an extra computational burden. A recent work~\citep{sauty2022progression} bridges the gap between the previous methods and novel deep learning techniques, using a Variational Autoencoder (VAE) to encode brain MRIs into a latent space, in which it defines a linear mixed effect model to learn the average disease trajectory from the population. Similarly, in~\citep{chadebec2022image}, the authors use a VAE to project images into a latent space and fit a generative model for disease progression with a fully variational approach. While population-based models provide advantages—such as an interpretable representation of average disease dynamics—they can become overly constrained when applied to high-dimensional data, where individual progression patterns vary significantly.

\subsection{Individual-based progression modeling}
Individual-based progression modeling is a more flexible framework in which models are directly trained to map current individual observations to their corresponding states at future timepoints. By operating at the individual-level, these models can use the subject's chronological age as the temporal axis, allowing for a better personalized modeling of disease trajectories. While individual-based models offer greater flexibility than population-based approaches, this often comes at the expense of interpretability regarding the underlying disease mechanisms. Many of these models leverage recent advances in deep generative techniques, making them particularly well-suited for handling high-dimensional data such as full MRI scans or 3D anatomical shapes. The most popular frameworks used for this task include Generative Adversarial Networks (GANs)~\citep{goodfellow2020generative}, VAEs~\citep{kingma2013auto}, Normalizing Flows (NFs)~\citep{papamakarios2021normalizing}, and most recently, diffusion models~\citep{ho2020denoising}. The next sections will describe these methods in greater detail.

\subsubsection{Generative adversarial networks}
GANs employ two competing neural networks — a generator that produces synthetic data and a discriminator that evaluates authenticity — to create increasingly realistic artificial outputs through iterative adversarial training. An example of this approach is proposed in~\citep{xia2021learning}. Here, the authors used a GAN-based method to simulate subject-specific brain aging trajectories, conditioned on the presence of AD, by minimizing adversarial and reconstruction losses without using longitudinal data. Among the first methods to model disease progression in 3D brain MRIs, the seminal work~\citep{ravi2022degenerative} introduces 4D-DaniNet, a generative model that exploits adversarial learning to provide individualized predictions of brain MRIs. To address memory limitations, 4D-DaniNet synthesizes 2D slices, which are subsequently reassembled into a 3D volume using a super-resolution module. 4D-DaniNet incorporates prior knowledge of disease progression by embedding biological constraints into the training loss function. In contrast, our proposed BrLP model introduces a key innovation by integrating prior knowledge directly into the image generation process, guiding synthesis at inference time. In~\citep{jung2021conditional}, the authors present ADESyn, a conditional GAN for synthesizing 3D brain MRIs across different stages of AD. The model generates 2D slices using an attention-based generator conditioned on disease progression and ensures 3D spatial consistency through integrated 2D and 3D discriminators. Instead of using the subject age as the temporal axis, ADESyn employs a disease condition score ranging from 0 (healthy) to 1 (AD), limiting its ability to distinguish changes in the early and late stages of AD. Lastly, CounterSynth~\citep{pombo2023equitable} is a GAN-based counterfactual synthesis method that can simulate various conditions within a brain MRI, including aging and disease progression. Rather than modeling the entire 3D brain MRI, CounterSynth generates a diffeomorphic transformation that warps the input image to reflect specified covariates (e.g., changes at a target age under a given condition). This technique preserves anatomical accuracy and minimizes artificial artifacts in the generated images.

\subsubsection{Variational autoencoders}
VAEs are deep generative models that learn to map data to and from a structured probabilistic latent space, enabling both compression and generation. In~\citep{he2024individualized}, the authors propose a double-encoder conditional VAE for predicting future MRI scans based on baseline MRI and subject metadata. The model also facilitates disease classification by estimating the posterior distribution when both baseline and follow-up MRIs are provided.

\subsubsection{Normalizing flows}
NFs use sequences of invertible transformations to convert simple distributions into complex ones while maintaining exact likelihood evaluation. Based on this, \citep{wilms2022invertible} introduces a bidirectional NF model that links brain morphology to age, potentially incorporating additional variables, such as disease diagnosis. This approach enables the generation of follow-up images at a target age and the estimation of brain age from a given image. Their model leverages NFs and, similar to CounterSynth~\citep{pombo2023equitable}, utilizes diffeomorphic deformations to represent structural changes.

\subsubsection{Diffusion models}\label{sec:diffusionmodels}
In recent years, diffusion models have emerged as a major breakthrough in generative AI, gaining significant attention for their state-of-the-art image synthesis capabilities and stable training processes, which avoid the challenges associated with adversarial approaches. A detailed overview of the diffusion model framework is provided in Section~\ref{sec:background}. These models have been successfully applied to 3D medical image synthesis, showing promising results~\citep{pinaya2022brain}. In the context of spatiotemporal modeling, the Sequence-Aware Diffusion Model (SADM)~\citep{yoon2023sadm} marks a significant advancement by allowing for the generation of longitudinal brain scans through autoregressive sampling informed by sequential MRI data. Other works have applied diffusion-based approaches to spatiotemporal modeling, but they are either limited to 2D slices of the brain~\citep{litrico2024tadm} or focused on different diseased organs~\citep{kim2022diffusion}. Building on recent advancements, our proposed BrLP method also uses a latent diffusion framework extended with a ControlNet~\citep{zhang2023adding} to more effectively model spatiotemporal dynamics, as detailed in the methodology section.

\subsection{Limitations of the existing methods}
In this section, we relate existing methodologies to the four key challenges outlined in Section~\ref{sec:introduction}. The first challenge pertains to the level of \textit{individualization} a model can achieve. Predictions from population-based progression models are not truly individualized, as the underlying model remains grounded in population-level patterns rather than being fully adapted to each patient's unique data. In contrast, individual-based models offer greater flexibility and personalization. Nevertheless, their potential is often underutilized, as many existing approaches~\citep{jung2021conditional,wilms2022invertible,yoon2023sadm} do not incorporate subject-specific metadata (e.g.,  demographic or diagnostic information) that could enhance individual-level predictions. The second challenge involves the effective exploitation of longitudinal data. Among the reviewed methods, only the work by~\citep{yoon2023sadm} explicitly leverages temporal sequences during inference, whereas the other approaches disregard this valuable information. The third challenge concerns \textit{spatiotemporal consistency}. Some earlier GAN-based approaches address this by enforcing identity preservation~\citep{xia2021learning,jung2021conditional} or incorporating biological constraints~\citep{ravi2022degenerative}. However, these solutions have seen diminishing attention, with more recent approaches~\citep{pombo2023equitable,he2024individualized,wilms2022invertible,yoon2023sadm} omitting explicit mechanisms to ensure temporally coherent and biologically plausible progression. The fourth and final challenge is the \textit{memory demand} associated with processing 3D medical images. Several methods address this by operating on 2D slices and subsequently stacking them to form a 3D volume~\citep{xia2021learning,ravi2022degenerative,jung2021conditional}. While computationally efficient, this strategy fails to capture inter-slice dependencies, which are critical for modeling inherently three-dimensional phenomena, highlighting the need for methodologies that operate directly on 3D images. Other approaches rely on lightweight deformation fields to model temporal progression as spatial transformations of a baseline image~\citep{wilms2022invertible,pombo2023equitable}. While deformation fields can preserve anatomical structures, they are inherently constrained to reshaping pre-existing anatomical content. As a result, they cannot generate new structures (e.g., white matter hypointensities in T1w images) not present in the initial scan. Another strategy is to rely on low-resolution volumes~\citep{he2024individualized}, which risks omitting clinically relevant anatomical details.\newline

Our proposed method, BrLP, addresses the limitations of existing approaches by harnessing the latest advances in diffusion modeling to deliver an individual-based framework that integrates subject-specific metadata, effectively utilizes available longitudinal data, enforces spatiotemporal consistency, and handles high-dimensional imaging data without sacrificing modeling capacity.

\section{Background - Diffusion models}\label{sec:background}
A Denoising Diffusion Probabilistic Model (DDPM)~\citep{ho2020denoising} is a deep generative model with two Markovian processes: forward diffusion and reverse diffusion. In the forward process, Gaussian noise is incrementally added to the original image $x_0$ over $T$ steps. At each step $t$, noise is introduced to the current image $x_{t-1}$ by sampling from a Gaussian transition probability defined as $q(x_t \mid x_{t-1}) \coloneqq \mathcal{N}(x_t; \sqrt{1 - \beta_t}x_{t-1}, \beta_t I)$, where $\beta_t$ follows a variance schedule. If $T$ is sufficiently large, $x_T$ will converge to pure Gaussian noise $x_T \sim \mathcal{N}(0, I)$. The reverse diffusion process aims to revert each diffusion step, allowing the generation of an image from the target distribution starting from pure noise $x_T$. The reverse transition probability has a Gaussian closed form, $q(x_{t-1} \mid x_t, x_0) = \mathcal{N}(x_{t-1} \mid \tilde\mu(x_0, x_t), \tilde\beta_t)$, conditioned on the real image $x_0$. As $x_0$ is not available during generation, a neural network is trained to approximate $\mu_\theta(x_t, t) \approx \tilde\mu(x_0, x_t)$. Following the work proposed in~\citep{ho2020denoising}, it is possible to reparameterize the mean in terms of $x_t$ and a noise term $\epsilon$, and then use a neural network to predict the noise $\epsilon_\theta(x_t, t) \approx \epsilon$, optimized with the following objective:

\begin{equation}
\mathcal{L}_{\epsilon} \coloneqq \mathbb{E}_{t, x_t, \epsilon \sim \mathcal{N}(0, I)} \left[ 
\lVert \epsilon - \epsilon_\theta(x_t, t) \rVert^2 \right].
\label{eqn:ddpmloss}
\end{equation}

An LDM~\citep{rombach2022high} extends the DDPM by applying the diffusion process to a latent representation $z$ of the image $x$, rather than to the image itself. This approach reduces the high memory demand while preserving the quality and flexibility of the models. The latent representation is obtained by training an autoencoder, composed of an encoder $\mathcal{E}$ and a decoder $\mathcal{D},$ such that the encoder maps the sample $x$ to the latent space $z = \mathcal{E}(x),$ and the decoder recovers it as $x = \mathcal{D}(z)$.

\begin{figure*}[t!]
    \centering
    \includegraphics[width=\textwidth]{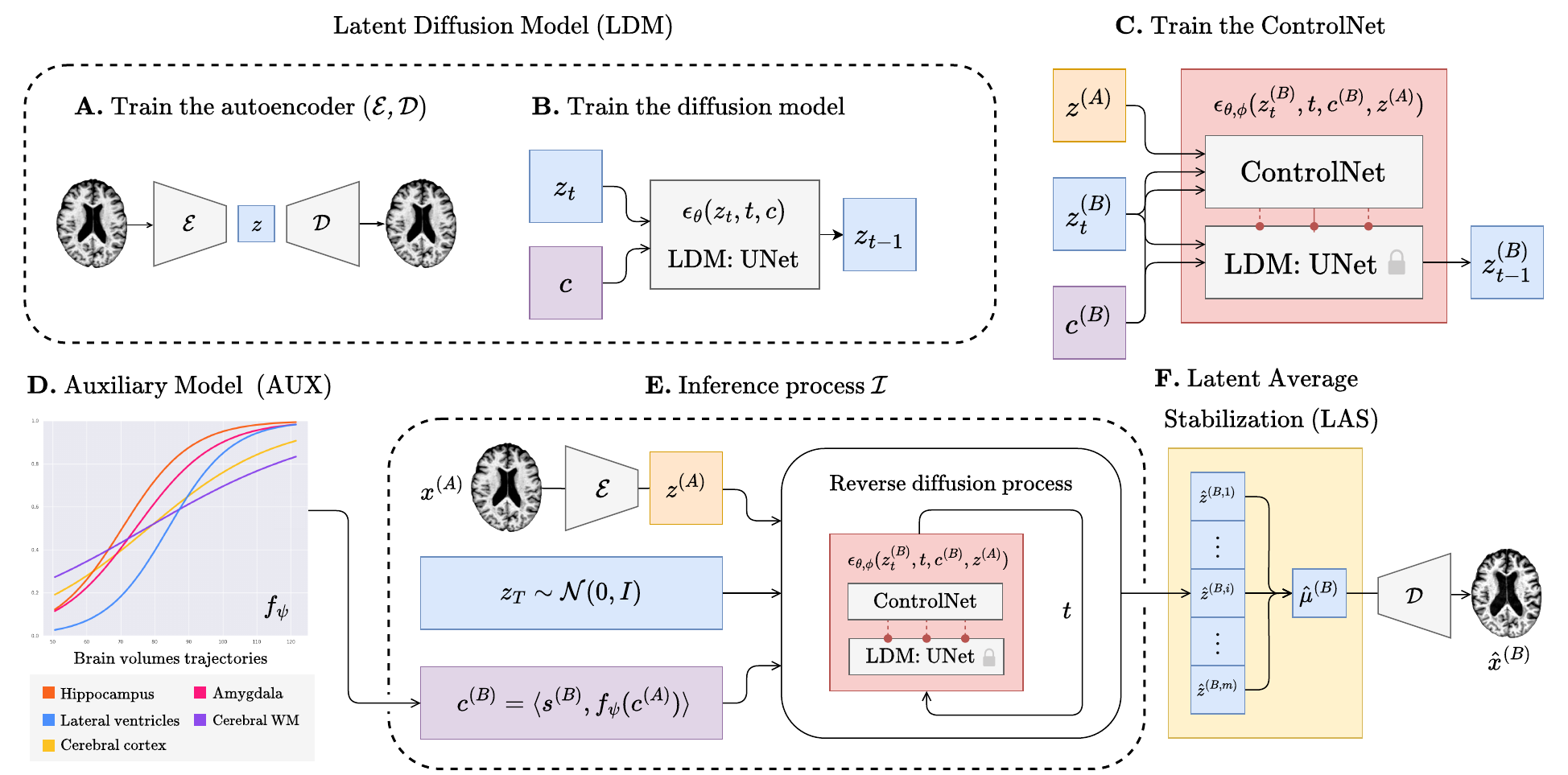}
    \caption{The overview of BrLP training and inference process. The training process outputs an autoencoder (A) that maps 3D brain MRIs into small latent representations; an LDM (B) able to generate latent representations according to subject-specific and progression-related covariates; a ControlNet (C), able to constrain the LDM’s generation process to a subject’s brain. During inference (E), progression-related variables at the target age are first predicted by an auxiliary model (D). These predictions, combined with subject-specific variables and the baseline MRI, condition the generation of the latent representations corresponding to the predicted brain at the target age. Finally, the LAS algorithm (F) repeats this process $m$ times and averages the obtained latent representations before decoding the result into the 3D MRI space.}
    \label{fig:pipeline}
\end{figure*}

\section{Methods - Brain Latent Progression (BrLP)}
We now introduce the architecture of BrLP, comprising four key components: an LDM, a ControlNet, an auxiliary model, and a LAS block, each described in successive paragraphs. These four components, summarized in Figure~\ref{fig:pipeline}, collectively address the challenges outlined in the introduction. In particular, the LDM is designed to generate random 3D brain MRIs that conform to specific covariates, while ControlNet aims to specialize these MRI scans to specific anatomical structures of a subject. Additionally, the auxiliary model leverages prior knowledge of disease progression to improve the precision in predicting the volumetric changes of specific brain regions. Finally, the LAS block is used during inference to improve spatiotemporal consistency, as well as to derive a measure of uncertainty for the predictions both at the global and voxel level.

\subsection{LDM - learning the brain MRI distribution}
Building upon~\citep{pinaya2022brain}, we train an LDM aimed to generate 3D brain MRIs mirroring specific covariates $c = \langle s, v \rangle$, where $s$ includes subject-specific metadata (age, sex, and cognitive status) while $v$ encompasses progression-related metrics such as volumes of brain regions (hippocampus, cerebral cortex, amygdala, cerebral white matter, and lateral ventricles) linked to AD progression~\citep{pini2016brain}. The construction of the LDM is a two-phase process. Initially, we train an autoencoder $(\mathcal{E}, \mathcal{D})$ (block A in Figure~\ref{fig:pipeline}) designed to produce a latent representation $z = \mathcal{E}(x)$ for each brain MRI $x$ within our dataset. Subsequently, we train a conditional UNet (block B in Figure~\ref{fig:pipeline}), represented as $\epsilon_\theta$, with network parameters $\theta$, aimed to estimate the noise $\epsilon_\theta(z_t, t, c)$ necessary for reverting from $z_t$ to $z_{t-1}$, as mentioned in Section~\ref{sec:diffusionmodels}. We train $\epsilon_\theta$ by minimizing the loss $\mathcal{L}_{\epsilon}$ (Eq.~\ref{eqn:ddpmloss}). Covariates $c$ are integrated into the network as conditions using a cross-attention mechanism, in line with~\citep{rombach2022high}. The generation process initiates by sampling random Gaussian noise $z_T \sim \mathcal{N}(0,I)$ and then iteratively reverses each diffusion step $z_t \to z_{t-1}$ for $t=T, \dots, 1$. Decoding the output $z_0$ from the final step $t=1$ yields a synthetic brain MRI $\hat x = \mathcal{D}(z_0)$ that follows the specified covariates $c$.

\subsection{ControlNet - conditioning on subject brain MRI}
The LDM provides only a limited degree of control over the generated brain MRI via the covariates $c$, and it does not allow for conditioning the model on individual anatomical structures. The purpose of this block is to extend the capabilities of the LDM to encompass this additional control. To achieve this, we use ControlNet~\citep{zhang2023adding}, (block C in Figure~\ref{fig:pipeline}) a neural network designed to work in conjunction with the LDM. We conceptualize ControlNet and LDM as a unified network $\epsilon_{\theta,\phi}$, where $\theta$ represents the fixed network's parameters of the LDM and $\phi$ denotes the trainable network's parameters of ControlNet. As in the LDM, $\epsilon_{\theta,\phi}$ is still used to predict the noise $\epsilon_{\theta,\phi}(z_t, t, c, z)$ in the reverse diffusion step $z_t \to z_{t-1}$, now incorporating $z = \mathcal E(x)$ as a condition to encompass the structure of the target brain $x$ during the generation process. To train ControlNet, we use the latent representations $z^{(A)}$ and $z^{(B)}$ from pairs of brain MRIs of the same patient taken at different ages $A$ < $B$. The ground-truth covariates $c^{(B)}$ associated with $z^{(B)}$ are known during training and used as target covariates. Each training iteration involves: i) sampling $t \sim U[1,T]$, ii) performing $t$ forward diffusion steps $z^{(B)} \to z_t^{(B)}$, iii) predicting the noise $\epsilon_{\theta,\phi}(z_t^{(B)}, t, c^{(B)}, z^{(A)})$ to revert $z^{(B)}_t \to z^{(B)}_{t-1}$, and iv) minimizing the loss $\mathcal{L}_{\epsilon}$ (Eq.~\ref{eqn:ddpmloss}).

\subsection{Proposed auxiliary model - leveraging disease prior knowledge}
\label{sec:auxiliarymodel}
AD-related regions shrink or expand over time and at different rates~\citep{pini2016brain}. Deep-learning-based spatiotemporal models strive to learn these progression rates directly from brain MRIs in a black-box manner, which can be very challenging. To aid this process, we propose incorporating prior knowledge of volumetric changes directly into our pipeline. To do so, we exploit an auxiliary model $f_\psi$ (block D in Figure~\ref{fig:pipeline}) able to predict how the volumes of AD-related regions change over time and provide this information to the LDM via the progression-related covariates $v$. The choice of our auxiliary model is tailored to two scenarios, making BrLP flexible for both cross-sectional and longitudinal data. For subjects with a single scan available at age $A$, we employ a regression model to estimate volumetric changes $\hat{v}^{(B)} = f_\psi(c^{(A)})$ at age $B$. For subjects with $n$ past visits accessible at ages $A_1, \dots, A_n$, we predict $\hat{v}^{(B)} = f_\psi(c^{(A_1)}, \dots, c^{(A_n)})$ using Disease Course Mapping (DCM)~\citep{schiratti2017bayesian,koval2021ad}, a model specifically designed for disease progression. DCM is intended to provide a more accurate trajectory in alignment with the subject's history of volumetric changes available. While we employ DCM as a potential solution, any suitable disease progression model can be used in BrLP.

\subsection{Inference process}
Let $x^{(A)}$ be the input brain MRI from a subject at age $A$, with known subject-specific metadata $s^{(A)}$ and progression-related volumes $v^{(A)}$ measured from $x^{(A)}$. As summarized in block E from Figure~\ref{fig:pipeline}, to infer the brain MRI $x^{(B)}$ at age $B > A$, we perform six steps: i) predict the progression-related volumes $\hat v^{(B)} = f_\psi(c^{(A)})$ using the auxiliary model; ii) concatenate this information with the subject-specific metadata $s^{(B)}$ to form the target covariates $c^{(B)} = \langle s^{(B)}, \hat v^{(B)}\rangle$; iii) compute the latent $z^{(A)} = \mathcal{E}(x^{(A)})$; iv) sample random Gaussian noise $z_T \sim \mathcal{N}(0,I)$; v) run the reverse diffusion process by predicting the noise $\epsilon_{\theta, \phi}(z_t, t, c^{(B)}, z^{(A)})$ to reverse each diffusion step for $t=T,\dots,1$; and finally vi) employ the decoder $\mathcal{D}$ to reconstruct the predicted brain MRI $\hat{x}^{(B)} = \mathcal{D}(z_0)$ in the imaging domain. This inference process is summarized into a compact notation $\hat z^{(B)} = \mathcal{I}(z_T, x^{(A)}, c^{(A)})$ and $\hat x^{(B)} = \mathcal{D}(\hat z^{(B)})$.

\subsection{Enhance inference via proposed Latent Average Stabilization (LAS)}
Variations in the initial value $x_T \sim \mathcal{N}(0,I)$ can lead to slight discrepancies in the results produced by the inference process. These discrepancies are especially noticeable when making predictions over successive timesteps, manifesting as irregular patterns or non-smooth transitions of progression. Therefore, we introduce LAS (block F in Figure~\ref{fig:pipeline}), a technique to improve spatiotemporal consistency by averaging different results of the inference process. In particular, LAS is based on the assumption that the predictions $\hat z^{(B)} = \mathcal{I}(z_T, x^{(A)}, c^{(A)})$ deviate from a theoretical mean $\mu^{(B)} = \mathbb{E}[\hat z^{(B)}]$. To estimate the expected value $\mu^{(B)}$, we propose to repeat the inference process $m$ times and average the results:

\begin{equation}
\mu^{(B)} =  \mathop{\mathbb{E}}_{z_T \sim \mathcal{N}(0,I)} \bigg[ \mathcal{I}(z_T, x^{(A)}, c^{(A)})\bigg] \approx \frac{1}{m}\sum^m \mathcal{I}(z_T, x^{(A)}, c^{(A)}).
\end{equation}

Similar to before, we decode the predicted scan as $\hat x^{(B)} = \mathcal{D}(\mu^{(B)})$.

\subsection{Quantifying the uncertainty of the prediction}
\label{sec:uncertainty}
Building on the assumption of the LAS algorithm, we interpret the spread of predictions around the theoretical mean as a measure of prediction uncertainty. Once the mean $\mu^{(B)}$ is approximated using LAS, we estimate the standard deviation of the predictions as follows:

\begin{equation}
    \sigma^{(B)} \approx \sqrt{\frac{\sum_{i=1}^m \left(z^{(B)}_i - \mu^{(B)}\right)^2}{m-1}}.
\end{equation}

We then average the components of $\sigma^{(B)}$ into a scalar global uncertainty measure, defined as $u^{(B)}$. 
To assess uncertainty at the voxel level, we decode the $m$ latent predictions into the 3D imaging domain and compute voxel-wise variance to derive an uncertainty map $U^{(B)}$:

\begin{equation}
    U^{(B)} = \frac{1}{m-1} \sum_{i=1}^m \left[ \mathcal{D}(z_i^{(B)}) - \mathcal{D}(\mu^{(B)}) \right]^2.
\end{equation}

\subsection{Implementation settings}
The architectures of the autoencoder, the UNet, and the ControlNet are taken from the Generative Models library~\citep{pinaya2023generative} available in MONAI~\citep{cardoso2022monai}. The LDM block, which includes the autoencoder $(\mathcal{E}, \mathcal{D})$ and the UNet $\epsilon_\theta$, follows the same settings proposed in~\citep{pinaya2022brain}. The autoencoder is fine-tuned from their available pre-trained model\footnotemark[1]. The training is performed using the Adam optimizer with a learning rate of $10^{-4}$ and a batch size of 8. The encoder $\mathcal{E}$ maps the input 3D brain MRI of shape $122 \times 146 \times 122$ into latent representations of shape $3 \times 16 \times 20 \times 16$.  The UNet $\epsilon_\theta$ is randomly initialized and trained using the AdamW optimizer with a learning rate of $2.5 \times 10^{-5}$ and a batch size of 16. During training, we use $T=1000$ and a scaled linear $\beta$ noise schedule from $\beta_1 = 0.0015$ to $\beta_T=0.0205$. The parameters $\phi$ of the ControlNet are randomly initialized and trained using the AdamW optimizer, with a learning rate of $2.5 \times 10^{-5}$ and a batch size of 16. The conditioning on the starting age $A$ is achieved by concatenating $A$ with the latent vector $z^{(A)}$ along the channel axis. The LDM's parameters $\theta$ remain fixed during this training phase. The settings for $T$ and the noise schedule $\beta_t$ are the same as those used in the LDM. The training is conducted on a GeForce RTX 4090 GPU with 24GB of VRAM. The training steps have required 51 hours for the autoencoder, 12 hours for the UNet, and 12 hours for the ControlNet, respectively. During inference, we use DDIM sampling~\citep{song2021denoising} with only 25 denoising steps. In Appendix~\ref{sec:ddim-steps}, we analyze the impact of using fewer denoising steps on BrLP's performance. For the LAS block, we parallelize the generation of multiple $\hat z^{(B)}$ on a single GPU, effectively reducing the overhead.

\footnotetext[1]{Model available at \url{https://github.com/Project-MONAI/GenerativeModels/tree/main/model-zoo/models/brain_image_synthesis_latent_diffusion_model}}

\subsection{MRI preprocessing}
Each T1w brain MRI used in our study is pre-processed using  N4 bias-field correction~\citep{tustison2010n4itk}, skull stripping~\citep{hoopes2022synthstrip}, affine registration to the MNI space, intensity normalization~\citep{shinohara2014statistical} and resampling to 1.5 mm\textsuperscript{3}. The volumes used as progression-related covariates and for our subsequent evaluation are calculated using SynthSeg 2.0~\citep{billot2023synthseg} and are expressed as percentages of the total brain volume to account for individual differences.


\section{Experiments and results}

In this section, we first describe the datasets and evaluation metrics used in our study. We then present an extensive evaluation of BrLP through five distinct experiments: an ablation study examining BrLP's components and hyperparameters, a comparative analysis against established baseline methods, an investigation of the impact of cognitive status conditioning, an assessment of our proposed uncertainty metrics at the global and voxel level, and an exploration of BrLP's potential to reduce Type II errors in clinical trials.

\begin{figure}[t!]
    \centering
    \includegraphics[width=\linewidth]{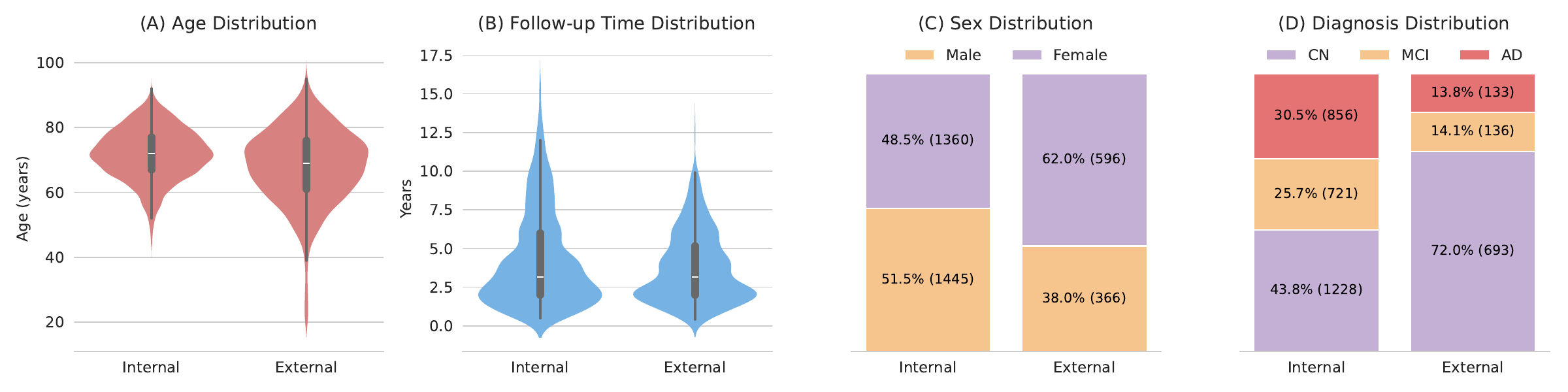}
    \caption{Demographic and diagnostic statistics of the internal and external datasets. Distributions include (A) age at baseline, (B) average time interval between the initial and follow-up visits, (C) sex distribution, and (D) diagnosis (CN, MCI, AD) at final visit.}
    \label{fig:dataset}
\end{figure}

\subsection{Internal and external datasets}


We collected a large internal dataset of T1w brain MRIs for training and internal evaluation, as well as an external dataset to further validate our findings on out-of-distribution data. For the internal dataset, we combined three publicly available longitudinal datasets: ADNI 1/2/3/GO (1,990 subjects)~\citep{petersen2010alzheimer}, OASIS-3 (573 subjects)~\citep{lamontagne2019oasis}, and AIBL (242 subjects)~\citep{ellis2009australian}. Dataset statements can be found in Appendix~\ref{sec:dataset-statements}. This internal dataset comprises 11,730 T1w brain MRI scans from 2,805 subjects, with each subject having at least two MRIs acquired during different visits.  We randomly split the dataset into a training set (80\%), a validation set (5\%), and a test set (15\%) with no overlapping subjects. The validation set is used for early stopping during training. For the external dataset, we used data from the NACC longitudinal study~\citep{beekly2007national}, including 2,257 T1w MRIs from 962 subjects used solely for evaluation. Figure~\ref{fig:dataset} provides an overview of the demographic and diagnostic statistics for both datasets.

\subsection{Evaluation metrics}
We evaluate BrLP using image-based and volumetric metrics to compare the predicted brain MRI scans with the subjects' actual follow-up scans. In particular, the Mean Squared Error (MSE) and the Structural Similarity Index (SSIM) are used to assess image similarity between the scans. Volumetric metrics in AD-related regions (hippocampus, amygdala, lateral ventricles, cerebrospinal fluid (CSF), and thalamus) evaluate the model's accuracy in tracking disease progression. Specifically, the Mean Absolute Error (MAE) between the volumes of actual follow-up scans and the generated brain MRIs is reported in the results. Notably, CSF and thalamus are excluded from progression-related covariates, enabling the analysis of unconditioned regions in our predictions.

\begin{table}[t!]
    \caption{Results from the ablation study. MAE (± SD) in predicted volumes is expressed as a percentage of total brain volume. The result is marked with a star if it is significantly better than all other configurations at the 5\% significance level (paired t-test with Bonferroni correction).}
    \label{tab:ablation}
    \setlength{\tabcolsep}{5pt}
    \def\arraystretch{1.5}
    \resizebox{\columnwidth}{!}{
    
    \begin{tabular}{l|c|cc|ccc|cc} \hline 
    & Exp. & \multicolumn{2}{c|}{\textbf{Image-based metrics}} & \multicolumn{3}{c|}{\textbf{MAE (conditional region volumes)}} & \multicolumn{2}{c}{\textbf{MAE (unconditional reg. volumes)}} \\
    \textbf{Method}& Settings & MSE $\downarrow$ & SSIM $\uparrow$ & Hippocampus $\downarrow$ & Amygdala $\downarrow$ & Lat. Ventricle $\downarrow$ & Thalamus $\downarrow$ & CSF $\downarrow$ \\
    \hline

    Base & - & 0.006 ± 0.003 &0.892 ± 0.030 &0.029 ± 0.026 &0.018 ± 0.016 &0.349 ± 0.415 &0.031 ± 0.023 &0.924 ± 0.718\\
    
    Base + AUX & DCM & 0.005 ± 0.002 &0.901 ± 0.027 &0.021 ± 0.018 &0.015 ± 0.012 &0.259 ± 0.267 &0.031 ± 0.025 &0.831 ± 0.656\\ 
    
    Base + LAS & - & 0.005 ± 0.003 &0.907 ± 0.030 &0.029 ± 0.025 &0.018 ± 0.016 &0.307 ± 0.396 &\textbf{0.029 ± 0.022} &0.902 ± 0.712\\
    
    Base + LAS + AUX & DCM & \textbf{0.004 ± 0.002}* &\textbf{0.914 ± 0.026}* &\textbf{0.020 ± 0.017}* &\textbf{0.014 ± 0.012}* &\textbf{0.231 ± 0.253}* &0.030 ± 0.024 &\textbf{0.799 ± 0.619}*\\
    
    \hline
     
    \end{tabular}
    }
\end{table}

\begin{figure}[t!]
    \centering
    \includegraphics[width=1\linewidth]{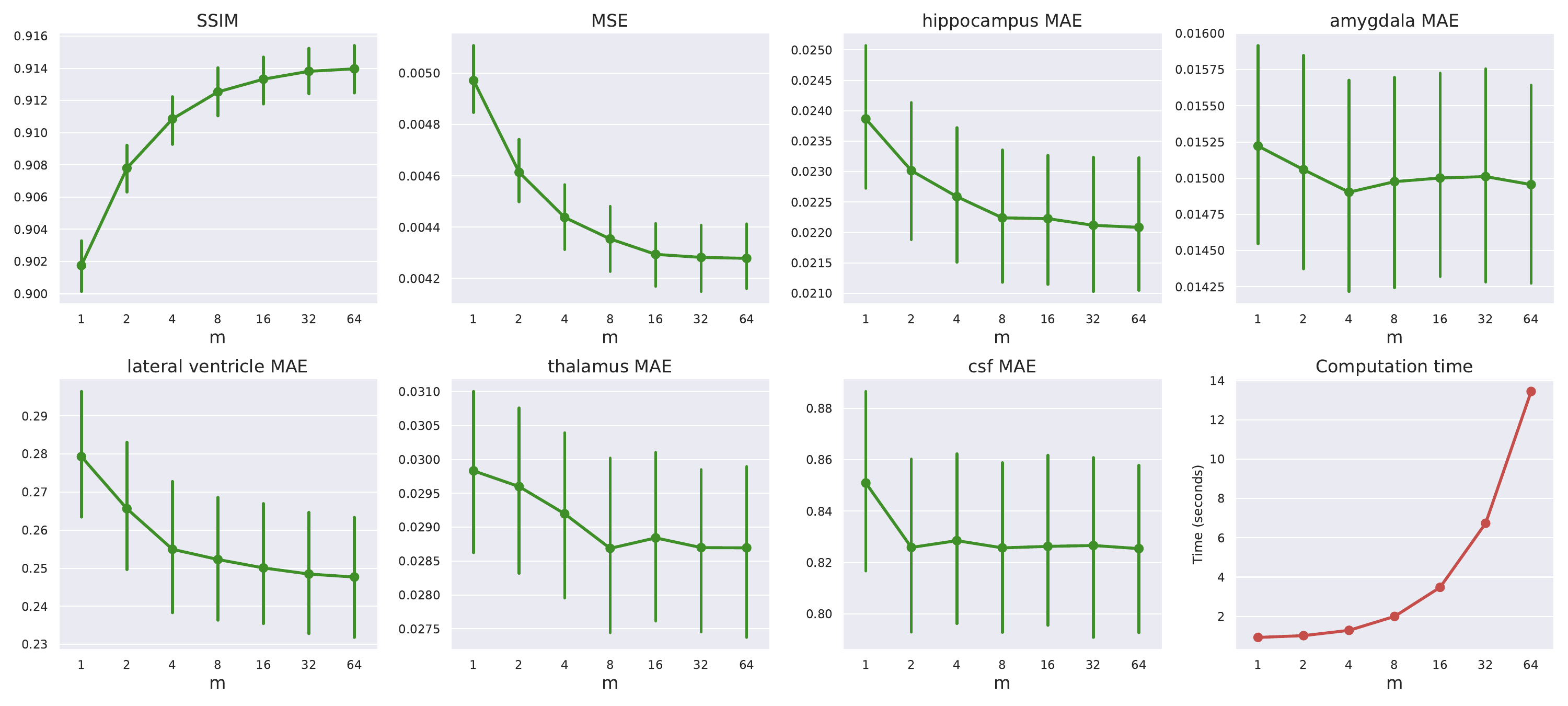}
    \caption{Effect of varying the LAS parameter $m$ on different performance metrics and computation time. The plots show the trends for SSIM, MSE, MAE for different brain regions (hippocampus, amygdala, lateral ventricle, thalamus, and CSF) and computation time as $m$ increases from 1 to 64. Error bars indicate the 95\% confidence intervals of the metric. Most metrics show improvements (higher SSIM, lower MSE and MAE) with increasing $m$.}
    \label{fig:ablation-on-m}
\end{figure}

\subsection{Ablation studies}\label{sec:ablation}
First, we evaluate how the LAS hyperparameter $m$ influences BrLP's predictive accuracy. As shown in Figure~\ref{fig:ablation-on-m}, our analysis across increasing values of $m$ ($m \in [1, 2, 4, 8, 16, 32, 64]$) reveals consistent improvements in terms of both image-based and volumetric metrics. In particular, increasing $m$ from 2 to 64 leads to gradual improvements: MSE decreases by 7\%, volumetric errors reduce by 3\% on average, and SSIM improves by 0.68\%. Paired t-tests ($\alpha = 0.05$) confirm that improvements are statistically significant for all metrics except for volumetric errors in the amygdala and CSF. However, this performance gain comes with increased computational demands. As $m$ increases, so does the computation time, as shown in Figure~\ref{fig:ablation-on-m}. Furthermore, since we perform the denoising of the $m$ latents in parallel on a single GPU, memory usage increases linearly with $m$: from 4.6 GB at $m=1$ to 9.6 GB at $m=64$ (excluding the memory required for VAE decoding). These results highlight the trade-off between accuracy and available computational resources. Unless otherwise specified, we set $m=64$ for all experiments from this point onward. \newline

Next, we conduct an ablation study to assess BrLP's performance with and without the auxiliary model (AUX) and the LAS. In this experiment, we used the first half of each subject's MRI visits to predict all subsequent MRIs in the latter half (see sequence-aware settings in Section~\ref{sec:comparison} for details). The results are presented in Table~\ref{tab:ablation}. The term ``Base'' in the table denotes BrLP without LAS (by setting $m{=}1$) and without AUX (i.e., no conditioning on progression-related covariates). The results show that both AUX and LAS independently improve prediction accuracy. Introducing the auxiliary model reduces volumetric error in conditioned regions by an average of 23\%, while its effect on unconditioned regions is smaller, yielding a 10\% reduction for CSF and no measurable improvement for the thalamus. These findings highlight the benefit of incorporating longitudinal information, and suggest that modeling all available regions with the auxiliary model could further enhance performance. LAS, on the other hand, contributes an additional average reduction of 4\% in volumetric error. When combined, AUX and LAS achieve the best overall performance, with a 21\% reduction in volumetric prediction error across all evaluated regions.

\begin{table}[t!]
    
    \caption{Results from the comparison with baseline methods on the internal test set. MAE (± SD) in predicted volumes is expressed as a percentage of total brain volume. The result is marked with a star if it is significantly better than all other methods at the 5\% significance level (paired t-test with Bonferroni correction).}
    \label{tab:quantitative-internal}
    
    \setlength{\tabcolsep}{5pt}
    \def\arraystretch{1.5}
    \resizebox{\columnwidth}{!}{
    
    \begin{tabular}{c|l|c|cc|ccc|cc} \hline 
    && & \multicolumn{2}{c|}{\textbf{Image-based metrics}} & \multicolumn{3}{c|}{\textbf{MAE (conditional region volumes)}} & \multicolumn{2}{c}{\textbf{MAE (unconditional reg. volumes)}} \\
    &\textbf{Method}& \textbf{Settings} & MSE $\downarrow$ & SSIM $\uparrow$ & Hippocampus $\downarrow$ & Amygdala $\downarrow$ & Lat. Ventricle $\downarrow$ & Thalamus $\downarrow$ & CSF $\downarrow$ \\

    \hline
    \multirow{5}{*}{\rotatebox[origin=c]{90}{\parbox[c]{2.5cm}{\centering \textbf{\color{Mulberry}All subjects\color{black}}}}}
    
    &DaniNet~\citep{ravi2022degenerative} & \multirow{3}{*}{\shortstack{Single\\ image}} & 0.016 ± 0.007 & 0.623 ± 0.162 & 0.030 ± 0.030 & 0.018 ± 0.017 & 0.257 ± 0.222 & 0.038  ± 0.030 & 1.081 ±  0.814 \\
    & CounterSynth~\citep{pombo2023equitable} & & 0.010 ± 0.004 &0.824 ± 0.052 &0.030 ± 0.018 &\textbf{0.014 ± 0.010} &0.310 ± 0.311 &0.127 ± 0.035 &0.881 ± 0.672\\
    & BrLP (proposed) & & \textbf{0.004 ± 0.002}* &\textbf{0.914 ± 0.028}* &\textbf{0.022 ± 0.021}* &0.015 ± 0.013 &\textbf{0.248 ± 0.300}* &\textbf{0.029 ± 0.023}* &\textbf{0.825 ± 0.623}*\\
    
    \cdashline{2-10}
    
    & Latent-SADM~\citep{yoon2023sadm} & \multirow{2}{*}{\shortstack{Sequence\\ aware}} & 0.008 ± 0.002 &0.855 ± 0.022 &0.035 ± 0.027 &0.018 ± 0.015 &0.329 ± 0.328 &0.037 ± 0.028 &0.924 ± 0.705\\
    & BrLP (proposed) & & \textbf{0.004 ± 0.002}* &\textbf{0.914 ± 0.026}* &\textbf{0.020 ± 0.017}* &\textbf{0.014 ± 0.012}* &\textbf{0.231 ± 0.253}* &\textbf{0.030 ± 0.024}* &\textbf{0.799 ± 0.619}*\\
    
    \hline
    \multirow{5}{*}{\rotatebox[origin=c]{90}{\parbox[c]{2.5cm}{\centering \textbf{\color{PineGreen}CN subjects only\color{black}}}}}
    
    &DaniNet~\citep{ravi2022degenerative} & \multirow{3}{*}{\shortstack{Single\\ image}} & 0.015 ± 0.007 & 0.662 ± 0.172 & 0.028 ± 0.028 & 0.017 ± 0.016 & 0.239 ± 0.206 & 0.035 ± 0.027 & 1.010 ± 0.761 \\
    & CounterSynth~\citep{pombo2023equitable} & & 0.011 ± 0.004 &0.807 ± 0.052 &0.035 ± 0.015 & 0.016 ± 0.009 &0.279 ± 0.194 &0.130 ± 0.033 &0.943 ± 0.614\\
    & BrLP (proposed) & & \textbf{0.004 ± 0.002}* &\textbf{0.916 ± 0.027}* &\textbf{0.017 ± 0.014}* & \textbf{0.011 ± 0.010}* &\textbf{0.175 ± 0.211}* &\textbf{0.028 ± 0.023}* &\textbf{0.807 ± 0.586}*\\
    
    \cdashline{2-10}
    
    & Latent-SADM~\citep{yoon2023sadm} & \multirow{2}{*}{\shortstack{Sequence\\ aware}} & 0.007 ± 0.002 &0.857 ± 0.021 &0.030 ± 0.021 &0.014 ± 0.011 &0.250 ± 0.228 &0.036 ± 0.028 &0.873 ± 0.647\\
    & BrLP (proposed) & & \textbf{0.004 ± 0.002}* &\textbf{0.916 ± 0.026}* &\textbf{0.017 ± 0.013}* &\textbf{0.011 ± 0.010}* &\textbf{0.165 ± 0.166}* &\textbf{0.031 ± 0.026}* &\textbf{0.816 ± 0.576}\\

    \hline
    \multirow{5}{*}{\rotatebox[origin=c]{90}{\parbox[c]{2.5cm}{\centering \textbf{\color{orange}MCI subjects only\color{black}}}}}
    
    &DaniNet~\citep{ravi2022degenerative} & \multirow{3}{*}{\shortstack{Single\\ image}} & 0.017 ± 0.007 & 0.652 ± 0.154 & 0.031 ± 0.029 & 0.018 ± 0.017 & 0.267 ± 0.213 & 0.040 ± 0.028 & 1.123 ± 0.782 \\
    & CounterSynth~\citep{pombo2023equitable} & & 0.010 ± 0.004 &0.836 ± 0.055 &0.031 ± 0.018 & 0.014 ± 0.010 &0.261 ± 0.200 &0.127 ± 0.037 & \textbf{0.906 ± 0.698}\\
    & BrLP (proposed) & & \textbf{0.004 ± 0.002}* &\textbf{0.915 ± 0.029}* &\textbf{0.019 ± 0.016}* & \textbf{0.013 ± 0.010} &\textbf{0.225 ± 0.220}* &\textbf{0.032 ± 0.025}* &\textbf0.916 ± 0.603\\

    \cdashline{2-10}
    
    & Latent-SADM~\citep{yoon2023sadm} & \multirow{2}{*}{\shortstack{Sequence\\ aware}} & 0.008 ± 0.002 &0.856 ± 0.023 &0.033 ± 0.022 &0.017 ± 0.013 &0.324 ± 0.316 &0.037 ± 0.028 &0.975 ± 0.702\\
    & BrLP (proposed) & & \textbf{0.004 ± 0.002}* &\textbf{0.918 ± 0.024}* &\textbf{0.019 ± 0.015}* &\textbf{0.013 ± 0.010}* &\textbf{0.233 ± 0.226}* &\textbf{0.033 ± 0.025}* &\textbf{0.873 ± 0.554}* \\

    \hline
    \multirow{5}{*}{\rotatebox[origin=c]{90}{\parbox[c]{2.5cm}{\centering \textbf{\color{red}AD subjects only\color{black}}}}}
    
    &DaniNet~\citep{ravi2022degenerative} & \multirow{3}{*}{\shortstack{Single\\ image}} & 0.017 ± 0.008 & 0.565 ± 0.147 & 0.033 ± 0.033 & 0.020 ± 0.019 & 0.283 ± 0.245 & 0.042 ± 0.033 & 1.184 ± 0.892 \\
    & CounterSynth~\citep{pombo2023equitable} & & 0.010 ± 0.003 &0.838 ± 0.042 & \textbf{0.024 ± 0.019}* &\textbf{0.012 ± 0.011}* &0.391 ± 0.461 &0.123 ± 0.035 &\textbf{0.776 ± 0.714}\\
    & BrLP (proposed) & & \textbf{0.005 ± 0.002}* &\textbf{0.910 ± 0.027}* & 0.031 ± 0.028 &0.021 ± 0.017 &\textbf{0.364 ± 0.403} &\textbf{0.027 ± 0.020}* & 0.782 ± 0.679\\
    
    \cdashline{2-10}
    
    & Latent-SADM~\citep{yoon2023sadm} & \multirow{2}{*}{\shortstack{Sequence\\ aware}} & 0.008 ± 0.002 &0.853 ± 0.023 &0.045 ± 0.035 &0.024 ± 0.018 &0.442 ± 0.410 &0.037 ± 0.027 &0.954 ± 0.776\\
    & BrLP (proposed) & & \textbf{0.005 ± 0.002}* &\textbf{0.910 ± 0.027}* &\textbf{0.026 ± 0.021}* &\textbf{0.020 ± 0.016}* &\textbf{0.323 ± 0.332}* &\textbf{0.027 ± 0.021}* &\textbf{0.718 ± 0.711}*\\

    \hline
    \end{tabular}
    }
\end{table}

\subsection{Quantitative and qualitative comparisons with baseline methods}\label{sec:comparison}

In this experiment, we compare our best BrLP setup with existing baseline methods. We categorize existing methods into single-image (cross-sectional) and sequence-aware (longitudinal) approaches. Single-image approaches, such as DaniNet and CounterSynth, predict progression using just one brain MRI as input. Sequence-aware methods, like SADM, leverage a series of prior brain MRIs as input. Due to the large memory demands of SADM, we have re-implemented it using an LDM, allowing the comparisons in our experiments. We refer to it as Latent-SADM. To evaluate all these methods, we conduct two separate experiments. In single-image methods, we predict all subsequent MRIs for a subject based on their initial scan. For sequence-aware methods, we use the first half of a subject's MRI visits to predict all subsequent MRIs in the latter half. In single-image settings, our approach uses an LM as the auxiliary model (detailed in Appendix~\ref{sec:app-aux-models-lm}). In contrast, for sequence-aware experiments, we employ the last available MRI in the sequence as the input for BrLP and fit a logistic DCM on the first half of the subject's visits as the auxiliary model (detailed in Appendix~\ref{sec:app-aux-models-dcm}). We conduct all experiments on both internal and external test sets. Table~\ref{tab:quantitative-internal} presents evaluation metrics for the internal test set, while Table~\ref{tab:quantitative-external} shows metrics for the external test set. Each table reports statistical significance for performance differences between BrLP and baseline methods, using a paired t-test with Bonferroni correction. We report results for the entire test set (all subjects) and further stratify them by cognitive status (CN subjects only, MCI subjects only, AD subjects only), allowing us to analyze potential prediction biases. \newline

On the internal test set, our method demonstrates a substantial improvement over baseline methods, achieving an average MSE reduction of 61.67\% (SD = 10.27\%) and an average SSIM increase of 21.51\% (SD = 17.89\%). For volumetric measurements across various brain regions, our approach outperforms the baselines, showing improvements of 18.84\% (SD = 10.27\%) over DaniNet, 24.61\% (SD = 28.64\%) over CounterSynth, and 25.46\% (SD = 10.17\%) over Latent-SADM. Similarly, on the external test set, BrLP maintains robust performance across both image-based and volumetric metrics, consistently surpassing baseline methods. Specifically, it achieves an average MSE reduction of 60.23\% (SD = 7.33\%) and an average SSIM increase of 22.84\% (SD = 18.41\%) relative to baselines. For volumetric measurements, improvements are observed at 17.60\% (SD = 7.21\%) over DaniNet, 25.74\% (SD = 25.08\%) over CounterSynth, and 25.91\% (SD = 9.47\%) over Latent-SADM. No notable differences appear between improvements in conditioned versus unconditioned regions in both the internal and external test sets. By examining the evaluation metrics in Table~\ref{tab:quantitative-internal} and Table~\ref{tab:quantitative-external} across different cognitive statuses, we identify a predictive bias in both BrLP and the baseline models, shown by performance differences among cognitive status groups. This bias is more pronounced in volumetric metrics: BrLP, DaniNet, and Latent-SADM show lower performance for AD, while CounterSynth underperforms for CN. These findings likely reflect that BrLP, DaniNet, and Latent-SADM underestimate the predicted progression, whereas CounterSynth overestimates the aging effects.\newline

\begin{table}[t!]
    
    \caption{Results from the comparison with baseline methods on the external test set. MAE (± SD) in predicted volumes is expressed as a percentage of total brain volume. The result is marked with a star if it is significantly better than all other methods at the 5\% significance level (paired t-test).}
    \label{tab:quantitative-external}
    
    \setlength{\tabcolsep}{5pt}
    \def\arraystretch{1.5}
    \resizebox{\columnwidth}{!}{
    
    \begin{tabular}{c|l|c|cc|ccc|cc} \hline 
    &&& \multicolumn{2}{c|}{\textbf{Image-based metrics}} & \multicolumn{3}{c|}{\textbf{MAE (conditional region volumes)}} & \multicolumn{2}{c}{\textbf{MAE (unconditional reg. volumes)}} \\
    &\textbf{Method}& \textbf{Settings} & MSE $\downarrow$ & SSIM $\uparrow$ & Hippocampus $\downarrow$ & Amygdala $\downarrow$ & Lat. Ventricle $\downarrow$ & Thalamus $\downarrow$ & CSF $\downarrow$ \\

    \hline
    \multirow{5}{*}{\rotatebox[origin=c]{90}{\parbox[c]{2.5cm}{\centering \textbf{\color{Mulberry}All subjects\color{black}}}}}
    
    &DaniNet~\citep{ravi2022degenerative} & \multirow{3}{*}{\shortstack{Single\\ image}} & 0.017 ± 0.007 & 0.611 ± 0.181 & 0.032 ± 0.031 & 0.018 ± 0.016 & 0.232 ± 0.210 & 0.039 ± 0.032 & 1.154 ± 0.871 \\
    & CounterSynth~\citep{pombo2023equitable} & & 0.011 ± 0.003 &0.813 ± 0.042 &0.030 ± 0.020 &\textbf{0.014 ± 0.010} &0.283 ± 0.314 &0.111 ± 0.034 &1.173 ± 0.731\\
    & BrLP (proposed) & & \textbf{0.005 ± 0.002}* &\textbf{0.909 ± 0.023}* &\textbf{0.024 ± 0.023}* & \textbf{0.014 ± 0.013}* &\textbf{0.213 ± 0.350}* &\textbf{0.030 ± 0.024}* &\textbf{1.044 ± 0.788}*\\
    
    \cdashline{2-10}
    
    & Latent-SADM~\citep{yoon2023sadm} & \multirow{2}{*}{\shortstack{Sequence\\ aware}} & 0.009 ± 0.002 &0.845 ± 0.020 &0.035 ± 0.029 &0.018 ± 0.015 &0.308 ± 0.349 &0.039 ± 0.030 &1.104 ± 0.858\\
    & BrLP (proposed) & & \textbf{0.004 ± 0.002}* &\textbf{0.912 ± 0.022}* &\textbf{0.022 ± 0.021}* &\textbf{0.013 ± 0.012}* &\textbf{0.209 ± 0.367}* &\textbf{0.030 ± 0.023}* &\textbf{1.000 ± 0.753}*\\
    
    \hline
    \multirow{5}{*}{\rotatebox[origin=c]{90}{\parbox[c]{2.5cm}{\centering \textbf{\color{PineGreen}CN subjects only\color{black}}}}}
    
    &DaniNet~\citep{ravi2022degenerative} & \multirow{3}{*}{\shortstack{Single\\ image}} & 0.016 ± 0.007 & 0.643 ± 0.190 & 0.031 ± 0.030 & 0.017 ± 0.016 & 0.225 ± 0.204 & 0.038 ± 0.031 & 1.108 ± 0.836 \\
    & CounterSynth~\citep{pombo2023equitable} & & 0.011 ± 0.003 &0.807 ± 0.040 &0.032 ± 0.019 & 0.015 ± 0.010 &0.271 ± 0.177 &0.113 ± 0.033 &1.311 ± 0.741\\
    & BrLP (proposed) & & \textbf{0.004 ± 0.002}* &\textbf{0.910 ± 0.023}* &\textbf{0.021 ± 0.018}* & \textbf{0.011 ± 0.010}* &\textbf{0.173 ± 0.186}* &\textbf{0.030 ± 0.024}* &\textbf{1.135 ± 0.815}*\\
    
    \cdashline{2-10}
    
    & Latent-SADM~\citep{yoon2023sadm} & \multirow{2}{*}{\shortstack{Sequence\\ aware}} & 0.009 ± 0.002 &0.846 ± 0.020 &0.030 ± 0.024 &0.015 ± 0.012 &0.258 ± 0.253 &0.038 ± 0.031 &1.187 ± 0.891\\
    & BrLP (proposed) & & \textbf{0.004 ± 0.002}* &\textbf{0.914 ± 0.020}* &\textbf{0.020 ± 0.017}* &\textbf{0.011 ± 0.010}* &\textbf{0.169 ± 0.180}* &\textbf{0.029 ± 0.023}* &\textbf{1.104 ± 0.780}*\\

    \hline
    \multirow{5}{*}{\rotatebox[origin=c]{90}{\parbox[c]{2.5cm}{\centering \textbf{\color{orange}MCI subjects only\color{black}}}}}

    &DaniNet~\citep{ravi2022degenerative} & \multirow{3}{*}{\shortstack{Single\\ image}} & 0.019 ± 0.006 & 0.576 ± 0.191 & 0.031 ± 0.032 & 0.020 ± 0.014 & 0.224 ± 0.217 & 0.037 ± 0.034 & 1.160 ± 0.866 \\
    & CounterSynth~\citep{pombo2023equitable} & & 0.010 ± 0.003 &0.831 ± 0.043 &\textbf{0.023 ± 0.016} &\textbf{0.012 ± 0.009} & \textbf{0.243 ± 0.212} &0.111 ± 0.034 &0.931 ± 0.592\\
    & BrLP (proposed) & & \textbf{0.005 ± 0.002}* &\textbf{0.907 ± 0.019}* & 0.029 ± 0.024 &0.015 ± 0.012 & 0.261 ± 0.222 &\textbf{0.029 ± 0.021}* &\textbf{0.870 ± 0.692}\\
    
    \cdashline{2-10}
    
    & Latent-SADM~\citep{yoon2023sadm} & \multirow{2}{*}{\shortstack{Sequence\\ aware}} & 0.009 ± 0.002 &0.845 ± 0.017 &0.040 ± 0.031 &0.019 ± 0.014 &0.356 ± 0.305 &0.039 ± 0.027 &0.899 ± 0.719\\
    & BrLP (proposed) & & \textbf{0.005 ± 0.002}* &\textbf{0.909 ± 0.018}* &\textbf{0.025 ± 0.019}* &\textbf{0.014 ± 0.012}* &\textbf{0.234 ± 0.174}* &\textbf{0.030 ± 0.021}* &\textbf{0.819 ± 0.664}*\\

    \hline
    \multirow{5}{*}{\rotatebox[origin=c]{90}{\parbox[c]{2.5cm}{\centering \textbf{\color{red}AD subjects only\color{black}}}}}
    
    &DaniNet~\citep{ravi2022degenerative} & \multirow{3}{*}{\shortstack{Single\\ image}} & 0.020 ± 0.008 & 0.481 ± 0.143 & 0.038 ± 0.037 & 0.021 ± 0.019 & 0.276 ± 0.250 & 0.046 ± 0.038 & 1.385 ± 1.045 \\
    & CounterSynth~\citep{pombo2023equitable} & & 0.010 ± 0.004 &0.827 ± 0.041 &\textbf{0.025 ± 0.028}* &\textbf{0.012 ± 0.012}* &0.383 ± 0.698 &0.101 ± 0.034 &\textbf{0.709 ± 0.526}\\
    & BrLP (proposed) & & \textbf{0.005 ± 0.003}* &\textbf{0.901 ± 0.028}* &0.036 ± 0.036 &0.025 ± 0.019 &\textbf{0.372 ± 0.781} &\textbf{0.031 ± 0.024}* & 0.758 ± 0.626\\
    
    \cdashline{2-10}
    
    & Latent-SADM~\citep{yoon2023sadm} & \multirow{2}{*}{\shortstack{Sequence\\ aware}} & 0.009 ± 0.002 &0.842 ± 0.023 &0.054 ± 0.042 &0.030 ± 0.023 &0.514 ± 0.626 &0.041 ± 0.032 &0.887 ± 0.730\\
    & BrLP (proposed) & & \textbf{0.005 ± 0.003}* &\textbf{0.904 ± 0.029}* &\textbf{0.030 ± 0.033}* &\textbf{0.021 ± 0.017}* &\textbf{0.380 ± 0.846}* &\textbf{0.032 ± 0.025}* &\textbf{0.671 ± 0.553}*\\
    
    \hline
    \end{tabular}
    }
\end{table}

\begin{figure*}[]
    \centering
    \includegraphics[width=0.7\linewidth]{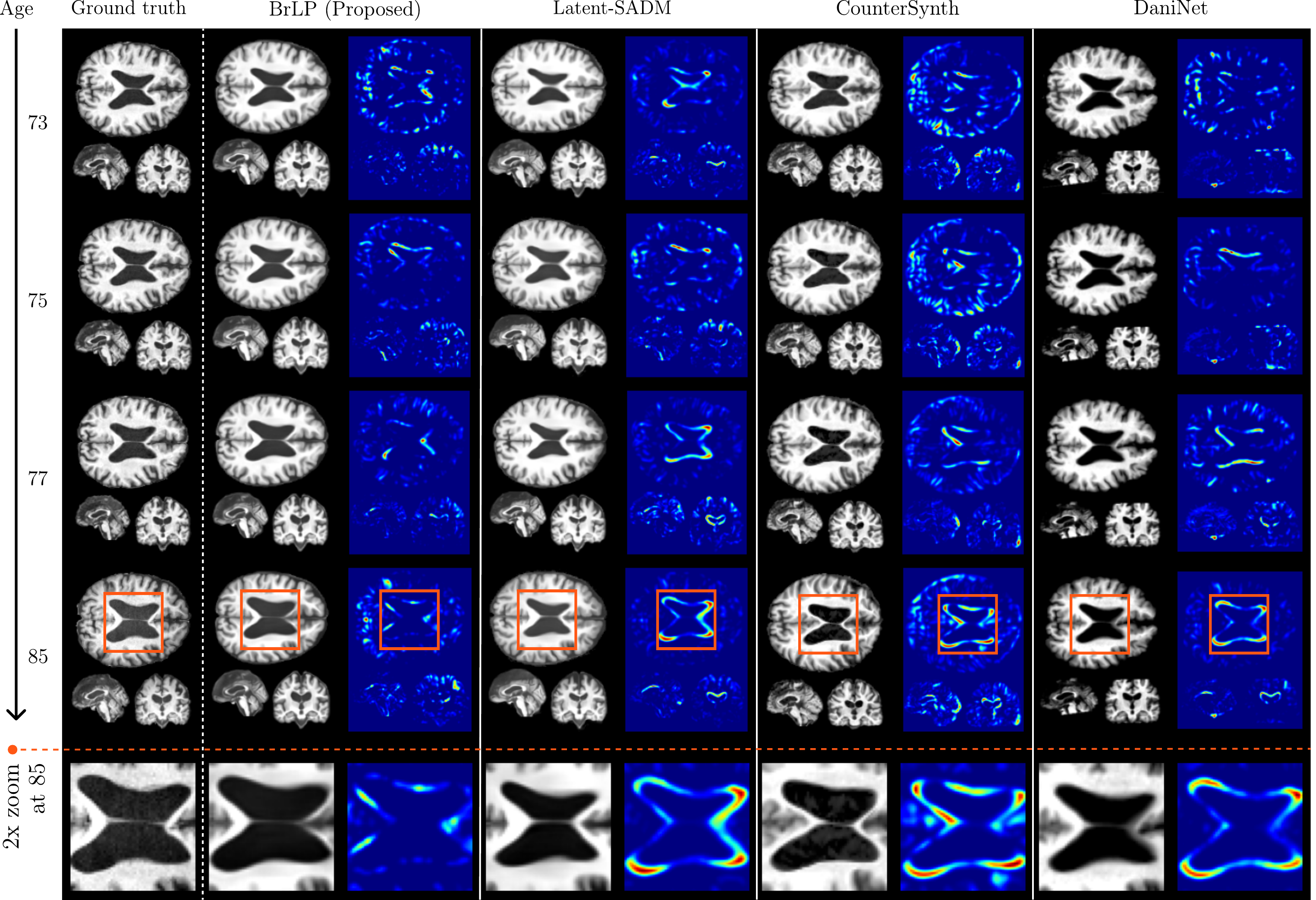}
    \caption{A comparison between the real progression of a 70 y.o. subject with MCI (from the internal test set) over 15 years and the predictions obtained by BrLP and the baseline methods. Each method shows a predicted MRI (left) and its deviation from the subject's real brain MRI (right).}
    \label{fig:comparison-internal}
\end{figure*}

Figure~\ref{fig:comparison-internal} provides a visual comparison between the actual progression of a 70-year-old subject with MCI from the internal dataset over a 15-year period and the predictions generated by BrLP and the baseline methods. The results from Latent-SADM and DaniNet exhibit a spatiotemporal mismatch in predicting the lateral ventricles' enlargement, whereas CounterSynth fails to capture the structural changes observed in the real progression. On the other hand, BrLP shows the most accurate prediction of the brain's anatomical changes, confirming the previous quantitative findings.

\begin{table}[t!]
    \caption{Evaluating the impact of incorrect conditioning on cognitive status in BrLP predictions. MAE (± SD) in predicted volumes is expressed as a percentage of total brain volume. The result is marked with a star if it is significantly better than all other configurations at the 5\% significance level (paired t-test with Bonferroni correction).}
    \label{tab:fakecn-experiment}
    \setlength{\tabcolsep}{5pt}
    \def\arraystretch{1.5}
    \resizebox{\columnwidth}{!}{
    
    \begin{tabular}{c|l|c|cc|ccc|cc} \hline 
    &&Exp. & \multicolumn{2}{c|}{\textbf{Image-based metrics}} & \multicolumn{3}{c|}{\textbf{MAE (conditional region volumes)}} & \multicolumn{2}{c}{\textbf{MAE (unconditional reg. volumes)}} \\
    &\textbf{Method}& Settings & MSE $\downarrow$ & SSIM $\uparrow$ & Hippocampus $\downarrow$ & Amygdala $\downarrow$ & Lat. Ventricle $\downarrow$ & Thalamus $\downarrow$ & CSF $\downarrow$ \\

    \hline
    \multirow{4}{*}{\rotatebox[origin=c]{90}{\parbox[c]{2cm}{\centering \textbf{\color{black}Internal\\test set\color{black}}}}}
    
    \multirow{2}{*} & Wrong cond. & \multirow{2}{*}{\shortstack{Single\\ image}}    
    & \textbf{0.005 ± 0.002} &\textbf{0.907 ± 0.027} &0.035 ± 0.029 &0.022 ± 0.018 &0.382 ± 0.455 &\textbf{0.027 ± 0.021} &0.833 ± 0.712\\
    & Correct cond. & & \textbf{0.005 ± 0.002} &\textbf{0.907 ± 0.027} &\textbf{0.031 ± 0.027}* &\textbf{0.021 ± 0.018}* &\textbf{0.372 ± 0.405} &0.028 ± 0.021 &\textbf{0.761 ± 0.677}*\\
    
    \cdashline{2-10}

    & Wrong cond. & \multirow{2}{*}{\shortstack{Sequence\\ aware}} & \textbf{0.005 ± 0.002} &\textbf{0.907 ± 0.027} &0.028 ± 0.023 &0.021 ± 0.016 &0.331 ± 0.356 &\textbf{0.028 ± 0.021} &0.753 ± 0.743\\
    & Correct cond. & & \textbf{0.005 ± 0.002} &\textbf{0.907 ± 0.027} &\textbf{0.026 ± 0.020} &\textbf{0.019 ± 0.016}* &\textbf{0.329 ± 0.338} &\textbf{0.028 ± 0.022} &\textbf{0.729 ± 0.720}\\

    \hline
    \multirow{4}{*}{\rotatebox[origin=c]{90}{\parbox[c]{2cm}{\centering \textbf{\color{black}External\\test set\color{black}}}}}
    
    & Wrong cond. & \multirow{2}{*}{\shortstack{Single\\ image}} & \textbf{0.005 ± 0.003} &\textbf{0.898 ± 0.029} &0.039 ± 0.037 &0.025 ± 0.020 &0.385 ± 0.790 &\textbf{0.031 ± 0.023} &0.795 ± 0.647\\
    & Correct cond. & & \textbf{0.005 ± 0.003} &\textbf{0.898 ± 0.029} &\textbf{0.036 ± 0.035}* &\textbf{0.024 ± 0.019} &\textbf{0.379 ± 0.762} &\textbf{0.031 ± 0.024} &\textbf{0.781 ± 0.621}\\

    \cdashline{2-10}

    & Wrong cond. & \multirow{2}{*}{\shortstack{Sequence\\ aware}} & \textbf{0.005 ± 0.003} &\textbf{0.901 ± 0.029} &0.034 ± 0.036 &0.023 ± 0.019 &\textbf{0.377 ± 0.855} &\textbf{0.032 ± 0.025} &0.692 ± 0.580\\
    & Correct cond. & & \textbf{0.005 ± 0.003} &\textbf{0.901 ± 0.029} &\textbf{0.030 ± 0.032} &\textbf{0.021 ± 0.017} &0.387 ± 0.853 &\textbf{0.032 ± 0.026} &\textbf{0.688 ± 0.548}\\

    \hline
    \end{tabular}
    }
\end{table}

\subsection{Evaluating the impact of incorrect conditioning on cognitive status in BrLP predictions}\label{sec:fakecn}
Due to the presence of a prediction bias towards healthy aging, we design an experiment to assess whether BrLP’s predictions are correctly influenced by altering the cognitive status of the input subject. Specifically, we conduct quantitative experiments where all AD subjects are conditioned as if they were CN, and we compare these results to those obtained with the correct conditioning. We perform this experiment both in single-image and sequence-aware settings and using both the internal and external test sets. We report the results for this experiment in Table~\ref{tab:fakecn-experiment}. As expected, our findings show that volumetric errors increase when incorrect cognitive status is provided, especially in the hippocampal region, which is significantly affected in AD subjects. While image-based metrics offer limited insight into the localized effects of conditioning, we observe clear and statistically significant differences in key regions such as the hippocampus in the single-image setting across both test sets. In contrast, these differences are not statistically significant in the sequence-aware setting. This suggests that the auxiliary model in the sequence-aware setting might leverage longitudinal information from prior visits to approximate the correct progression rate of AD subjects, thereby mitigating the impact of incorrect conditioning.

\begin{figure*}[t!]
    \centering
    \includegraphics[width=1\linewidth]{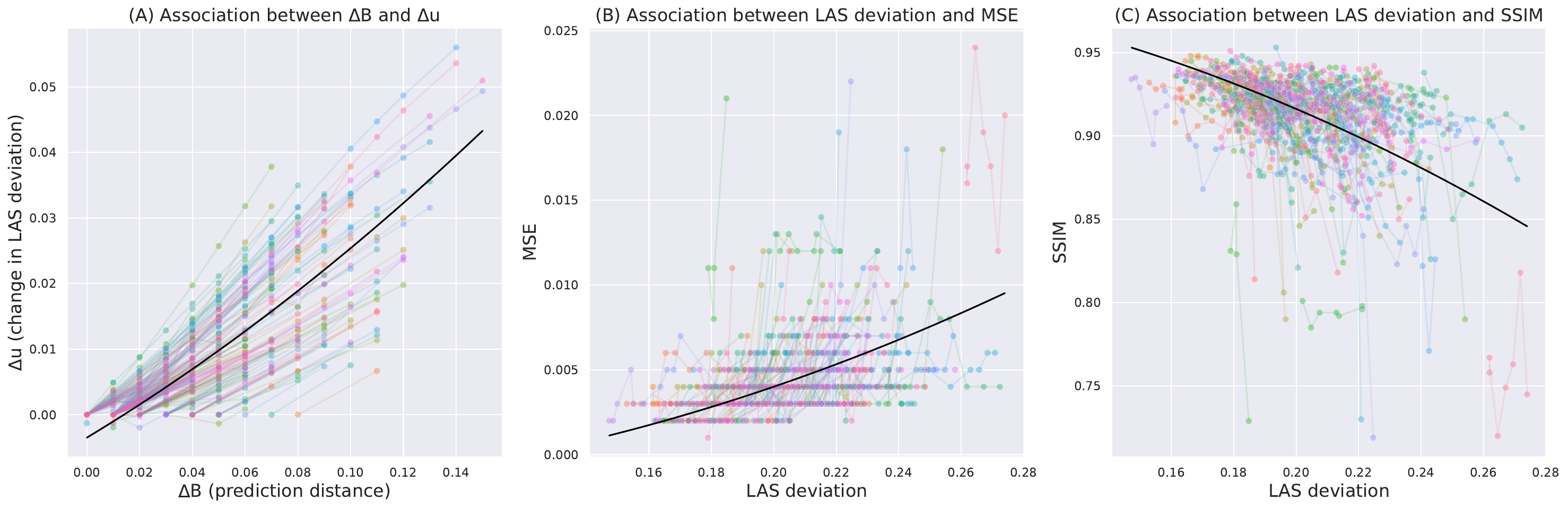}
    \caption{(A) Difference in uncertainty ($y$-axis) as a function of prediction distance ($x$-axis) in years (divided by 100). (B) MSE ($y$-axis) as a function of uncertainty ($x$-axis). (C) SSIM ($y$-axis) as a function of uncertainty ($x$-axis). In all plots, colored lines represent trends for individual subjects, and the black line shows the overall fixed effect from a linear mixed-effects model.}
    \label{fig:uncertainty-analysis}
\end{figure*}

\begin{figure}[t!]
    \centering
    \includegraphics[width=1\linewidth]{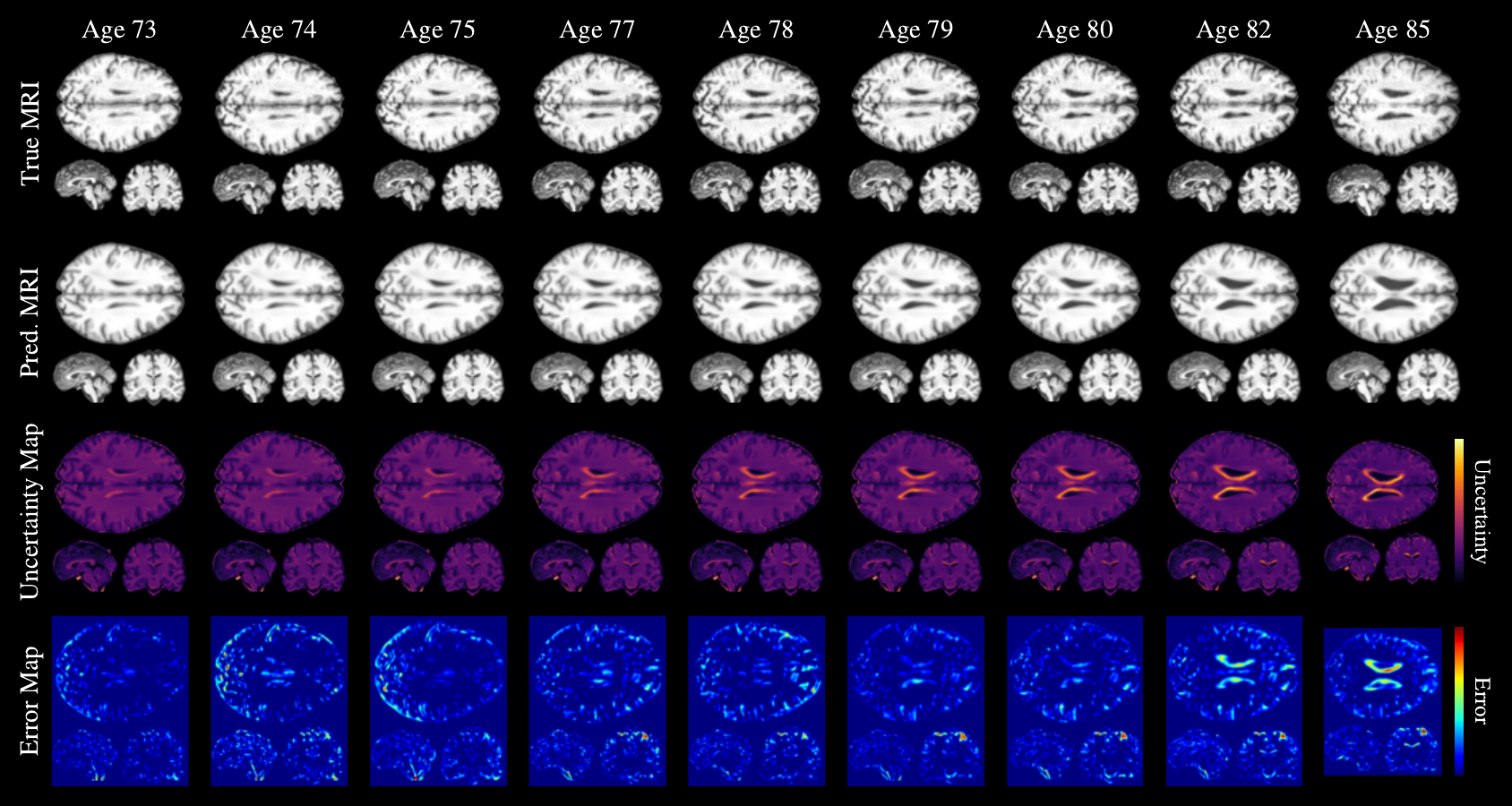}
    \caption{Voxel-level uncertainty evaluated for predictions at different timesteps for a single subject. The first and second rows show the ground truth and predicted MRIs, respectively, at each timestep. The third row presents the uncertainty maps, with lighter colors indicating higher uncertainty. The fourth row displays the voxel-level squared error between the ground truth and predicted MRIs.}
    \label{fig:uncertainty-map}
\end{figure}

\subsection{Analysis of the proposed uncertainty measures}\label{sec:lasdeviation}
In the next three experiments we evaluate the suitability of the uncertainty measures proposed in Section~\ref{sec:uncertainty}. First, we examine the relationship between global-level uncertainty and prediction distance, defined as the temporal gap between the age at input MRI and the target age. Intuitively, uncertainty should increase with prediction distance due to the considerable temporal heterogeneity inherent in disease progression. Second, we analyze the relationship between global-level uncertainty and prediction error using image-based metrics. Also in this case, we expect the error to be higher in predictions with higher uncertainty. We use linear mixed-effects models to investigate on these relationships, and we report the fixed effect coefficient $\beta$ along with the p-value $p$. In the third experiment, we evaluate the voxel-level uncertainty map quantitatively by measuring its voxel-wise correlation with prediction error, and qualitatively by visualizing how the map evolves with increasing prediction distance.

\subsubsection{Global-level uncertainty increases with prediction distance}\label{sec:lasdeviation-dist}
Here, we analyze the relationship between prediction distance and global-level uncertainty. For each predicted follow-up MRI, the prediction distance ($\Delta B_i$) is calculated as the difference between the follow-up age ($B_i$) and the starting age ($A$). The final uncertainty difference ($\Delta u_i$) is computed as the difference between the uncertainty at the $i$-th prediction ($u_i$) and the uncertainty at the first prediction ($u_1$). We use $\Delta B_i$ and $\Delta B_i^2$ as independent variables to predict $\Delta u_i$, with subject ID as a random effect. We find that the global-level uncertainty significantly increases with prediction distance $\Delta B_i$ ($\beta$ = 0.243, \textit{p} < 0.001) and $\Delta B_i^2$ ($\beta$ = 0.460, \textit{p} < 0.001). Figure~\ref{fig:uncertainty-analysis}-A shows the fixed effects from the mixed-effects model, along with the observed values of these variables for individual subjects.

\subsubsection{Global-level uncertainty associates with prediction error}\label{sec:lasdeviation-error}
Here, we analyze the relationship between the proposed global-level uncertainty measure and image-based metrics (MSE and SSIM). Specifically, we use the square of the uncertainty ($u^2$) as the independent variable to predict the image-based metrics, while incorporating subject ID as a random effect. Our results demonstrate a significant positive correlation between MSE and $u^2$ ($\beta = 0.157$, \textit{p} < 0.001), indicating that higher uncertainty corresponds to increased mean squared error. At the same time, we find that SSIM exhibits a significant negative correlation with $u^2$ ($\beta = -2.008$, \textit{p} < 0.001), suggesting that higher global-level uncertainty is associated with decreased structural similarity. We illustrate these relationships in Figure~\ref{fig:uncertainty-analysis}-B and Figure~\ref{fig:uncertainty-analysis}-C, which present both the fixed effects estimated by our model and the observed values from individual subjects. These findings collectively support the utility of our global-level uncertainty measure as a predictor of image quality metrics in the context of BrLP predictions.

\subsubsection{Evaluating the voxel-level uncertainty map}
In this section, we evaluate the voxel-level uncertainty map using both quantitative and qualitative analyses. For the quantitative assessment, we adopt the methodology of~\citep{zhou2024cascaded} and compute the correlation between the uncertainty map $U^{(B)}$ and voxel-wise squared prediction error (between predicted and ground-truth follow-up images). We explicitly exclude background voxels to prevent artificial inflation of the correlation. We obtain a Spearman correlation coefficient of $0.63 \pm 0.11$, suggesting that the model assigns higher uncertainty to regions with greater prediction error. In Figure~\ref{fig:uncertainty-map}, we showcase a qualitative example of the obtained voxel-level map, illustrating how uncertainty evolves with increasing prediction distance. We observe a gradual rise in uncertainty around the lateral ventricles, consistent with the findings reported in Section~\ref{sec:lasdeviation-dist}. Moreover, the uncertainty map visually corresponds with the error map in the latest predictions, reinforcing the quantitative results. Together, these findings suggest that the model's uncertainty estimates are both meaningful and informative for assessing prediction reliability at the voxel level.

\begin{figure}[t!]
    \centering
    \includegraphics[width=1\linewidth]{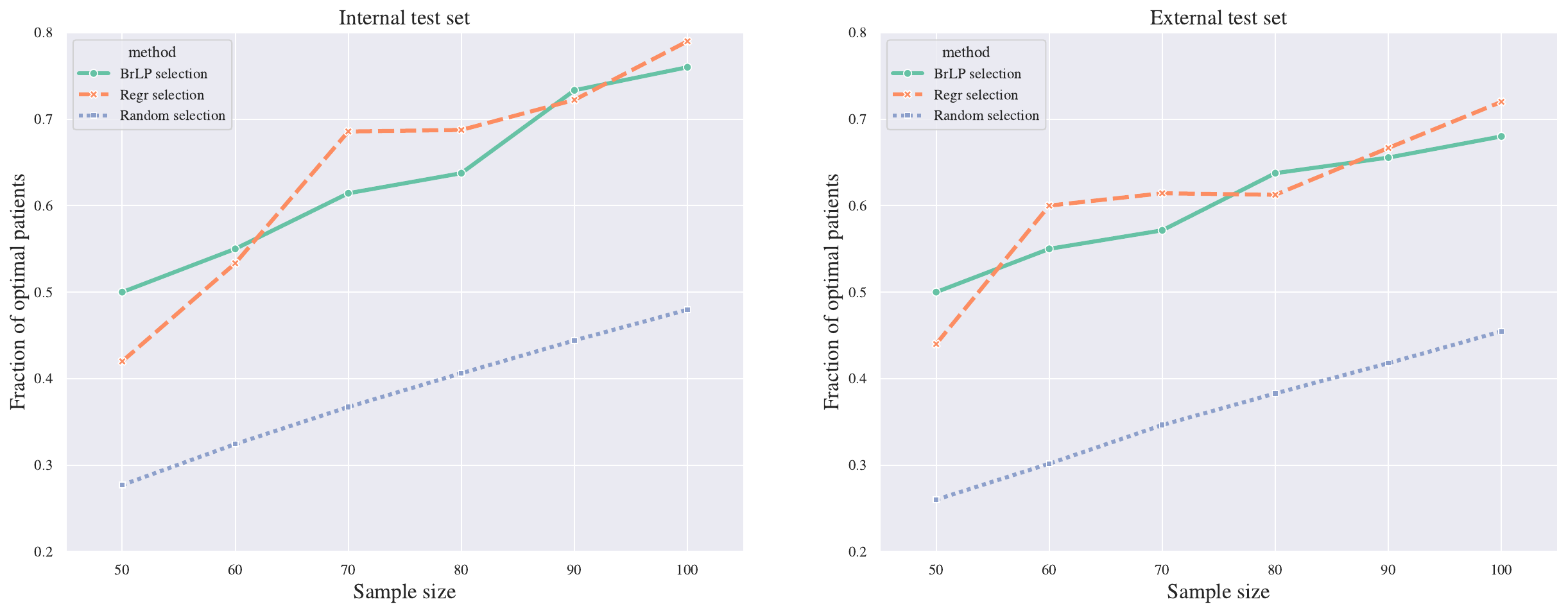}
    \caption{Comparison of patient selection methods for identifying fast progressors in clinical trials. The plot shows the efficacy ($y$-axis) of three selection methods (Random, BrLP, and Regression) across various sample sizes ($x$-axis) in both internal and external test sets. Efficacy is measured as the proportion of fast progressors (based on hippocampal atrophy) correctly identified by each method compared to the optimal selection.}
    \label{fig:patient-selection}
\end{figure}

\subsection{Downstream application: avoiding type II errors in clinical trials using BrLP}\label{sec:application}
Type II errors in clinical trials occur when a study fails to demonstrate the efficacy of a treatment that actually has significant effects. This can result in substantial financial losses and the failure to bring potentially beneficial drugs to market. One contributing factor to Type II errors is the inclusion of patients whose disease progression is too slow to observe treatment effects within the trial duration. Consequently, there is growing interest in identifying and selecting "fast progressors" - subjects whose disease advances more rapidly~\citep{wang2024multimodal,birkenbihl2024deep}. Predictive tools such as BrLP  can be used for this downstream application, providing predictive capabilities to assess disease progression at an individual level and facilitating the identification of fast progressors. \newline

We evaluate BrLP's ability to identify these fast progressors by designing a retrospective study involving all subjects with two-year follow-up MRI data from both internal (154 subjects) and external (165 subjects) datasets. Ground truth is established by ranking subjects based on the largest observed hippocampal atrophy, quantified as the relative reduction in hippocampal volume. We compare BrLP to a standard regression-based strategy (detailed in Section~\ref{sec:app-aux-models-lm}). Specifically, BrLP predicts each subject's two-year follow-up MRI to estimate atrophy rates based on the predicted hippocampal volumes. In contrast, the regression-based approach directly predicts two-year hippocampal volumes from baseline data without generating full MRI scans. For both methods, the top $S$ candidates are selected based on the largest predicted changes in hippocampal volume. To simulate different trial sizes, we tested $S \in [50, 60, 70, 80, 90, 100]$ and evaluated selection performance by calculating the overlap between the selected candidates and the ground truth rankings. \newline

Figure~\ref{fig:patient-selection} presents the results of these experiments. We report performance as improvements over a random selection strategy, averaged across all values of $S$. The regression model achieves the highest performance, selecting 25.67\% more fast progressors than random selection on the internal test set, and 24.83\% more on the external test (representing a 0.84\% drop in performance). BrLP achieves similar results, identifying 24.94\% more fast progressors than random selection internally and 23.85\% externally (a 1.09\% drop in performance). Although BrLP was not explicitly trained for this task, its performance is comparable to that of the regression model, highlighting its potential for downstream clinical applications such as patient selection for clinical trials.

\section{Discussion}

In this work, we propose BrLP, a novel individual-based spatiotemporal model that leverages recent advances in diffusion models to address the limitations of existing methodologies. Our approach is designed to work with both cross-sectional and longitudinal data inputs. By utilizing compact latent representations derived from full 3D brain MRIs, BrLP significantly reduces memory requirements, allowing it to operate on consumer-grade GPUs. This efficiency makes the method accessible for use in cost-sensitive environments, such as hospitals and research centers. \newline

In Section~\ref{sec:fakecn}, we demonstrate that BrLP effectively exploits the conditioning covariates, yielding better quantitative results when using the correct cognitive status of each test subject. In Section~\ref{sec:ablation}, we show that incorporating the auxiliary model enhances BrLP's performance, highlighting the importance of exploiting longitudinal data when developing spatiotemporal models. Additionally, we illustrate in the same section that implementing the LAS algorithm improves overall performance, though it introduces a trade-off between computational overhead—due to the hyperparameter $m$—and prediction accuracy. Section~\ref{sec:lasdeviation} demonstrates that the global- and voxel-level uncertainty measures derived from the LAS algorithm are associated with both prediction distance and prediction error, offering a potential tool for assessing prediction reliability in clinical applications. Finally, in Section~\ref{sec:application}, we showcase how BrLP predictions can identify patients with faster disease progression during clinical trials, potentially reducing the risk of type II errors. \newline

While our results highlight BrLP’s strengths, some experiments also reveal a few limitations. The first limitation is a bias in predictive performance: quantitative metrics are generally less accurate for AD cases compared to healthy aging (see Section~\ref{sec:comparison}). This discrepancy is likely due to the inherent heterogeneity and complexity of AD progression. Interestingly, a similar bias is observed across all baseline models, suggesting that it is not unique to BrLP. The second limitation involves a possible sex-related bias: as shown in Appendix~\ref{sec:sex-differences}, the model performs slightly better on female subjects in terms of image-based metrics. A third limitation, as also highlighted in~\citep{mcmaster2025technical}, is a smoothing effect in BrLP’s outputs. This effect is likely introduced by the VAE component~\citep{rivera2024train} and may reduce the model’s ability to capture fine-grained details in the data.\newline

Future research will aim to reduce the performance gap between the healthy and AD predictions by further disentangling disease progression from the ageing process and incorporating additional disease-specific variables that influence the severity and pace of progression in individuals. One potential direction is to model disease stage transitions (e.g., from MCI to AD) during longitudinal predictions, rather than assuming the subject’s final cognitive status.  Finally, adapting and applying BrLP to other brain diseases and chronic conditions affecting various organs will further demonstrate its versatility and clinical impact.

\section{Conclusion}
This work introduces BrLP, a 3D spatiotemporal model that accurately captures neurodegenerative disease progression patterns by predicting individual 3D brain MRI evolution. While we focused on brain MRI applications, BrLP’s potential extends to other imaging modalities and progressive diseases. Moreover, the model can potentially incorporate additional covariates, such as genetic data, for enhanced individualization. Our experiments demonstrate how BrLP can be used for patient selection in clinical trials to reduce the risk of type II errors. We believe that its application also extends to post-trial analysis, where, by generating digital twins of participants, BrLP could simulate untreated disease trajectories, enabling individualized treatment effect assessment. This approach could reduce the reliance on control groups and mitigate ethical concerns related to withholding potential therapeutic benefits.


\section*{Data availability}
The link to the code is provided in the manuscript. Access to the ADNI and AIBL datasets can be requested through the LONI IDA portal (\url{https://ida.loni.usc.edu/}). The OASIS-3 dataset is available upon request at \url{https://sites.wustl.edu/oasisbrains/}, while access to the NACC dataset can be requested via \url{https://naccdata.org/}.

\section*{Acknowledgments}

This work has been partially supported by the Italian Ministry of Health, Piano Operativo Salute (POS) trajectory 2 “eHealth, diagnostica avanzata, medical device e mini invasivita” through the project “Rete eHealth: AI e strumenti ICT Innovativi orientati alla Diagnostica Digitale (rAIdD)” (CUP J43C22000380001). L. Puglisi is enrolled in a PhD program at the University of Catania, fully funded by PNRR (DM 118/2023). \newline

\textit{Data were provided in part by OASIS Longitudinal Multimodal Neuroimaging: Principal Investigators: T. Benzinger, D. Marcus, J. Morris; NIH P30 AG066444, P50 AG00561, P30 NS09857781, P01 AG026276, P01 AG003991, R01 AG043434, UL1 TR000448, R01 EB009352. AV-45 doses were provided by Avid Radiopharmaceuticals, a wholly owned subsidiary of Eli Lilly.}\newline

\textit{Data collection and sharing for the Alzheimer's Disease Neuroimaging Initiative (ADNI) is funded by the National Institute on Aging (National Institutes of Health Grant U19AG024904). The grantee organization is the Northern California Institute for Research and Education. In the past, ADNI has also received funding from the National Institute of Biomedical Imaging and Bioengineering, the Canadian Institutes of Health Research, and private sector contributions through the Foundation for the National Institutes of  Health (FNIH) including generous contributions from the following: AbbVie, Alzheimer’s Association; Alzheimer’s Drug Discovery Foundation; Araclon Biotech; BioClinica, Inc.; Biogen; Bristol-Myers Squibb Company; CereSpir, Inc.; Cogstate; Eisai Inc.; Elan Pharmaceuticals, Inc.; Eli Lilly and Company; EuroImmun; F. Hoffmann-La Roche Ltd and its affiliated company Genentech, Inc.; Fujirebio; GE Healthcare; IXICO Ltd.; Janssen Alzheimer Immunotherapy Research \& Development, LLC.; Johnson \& Johnson Pharmaceutical Research \& Development LLC.; Lumosity; Lundbeck; Merck \& Co., Inc.; Meso Scale Diagnostics, LLC.; NeuroRx Research; Neurotrack Technologies; Novartis Pharmaceuticals Corporation; Pfizer Inc.; Piramal Imaging; Servier; Takeda Pharmaceutical Company; and Transition Therapeutics} \newline

\textit{The NACC database is funded by NIA/NIH Grant U24 AG072122. NACC data are contributed by the NIA-funded ADRCs: P30 AG062429 (PI James Brewer, MD, PhD), P30 AG066468 (PI Oscar Lopez, MD), P30 AG062421 (PI Bradley Hyman, MD, PhD), P30 AG066509 (PI Thomas Grabowski, MD), P30 AG066514 (PI Mary Sano, PhD), P30 AG066530 (PI Helena Chui, MD), P30 AG066507 (PI Marilyn Albert, PhD), P30 AG066444 (PI John Morris, MD), P30 AG066518 (PI Jeffrey Kaye, MD), P30 AG066512 (PI Thomas Wisniewski, MD), P30 AG066462 (PI Scott Small, MD), P30 AG072979 (PI David Wolk, MD), P30 AG072972 (PI Charles DeCarli, MD), P30 AG072976 (PI Andrew Saykin, PsyD), P30 AG072975 (PI David Bennett, MD), P30 AG072978 (PI Neil Kowall, MD), P30 AG072977 (PI Robert Vassar, PhD), P30 AG066519 (PI Frank LaFerla, PhD), P30 AG062677 (PI Ronald Petersen, MD, PhD), P30 AG079280 (PI Eric Reiman, MD), P30 AG062422 (PI Gil Rabinovici, MD), P30 AG066511 (PI Allan Levey, MD, PhD), P30 AG072946 (PI Linda Van Eldik, PhD), P30 AG062715 (PI Sanjay Asthana, MD, FRCP), P30 AG072973 (PI Russell Swerdlow, MD), P30 AG066506 (PI Todd Golde, MD, PhD), P30 AG066508 (PI Stephen Strittmatter, MD, PhD), P30 AG066515 (PI Victor Henderson, MD, MS), P30 AG072947 (PI Suzanne Craft, PhD), P30 AG072931 (PI Henry Paulson, MD, PhD), P30 AG066546 (PI Sudha Seshadri, MD), P20 AG068024 (PI Erik Roberson, MD, PhD), P20 AG068053 (PI Justin Miller, PhD), P20 AG068077 (PI Gary Rosenberg, MD), P20 AG068082 (PI Angela Jefferson, PhD), P30 AG072958 (PI Heather Whitson, MD), P30 AG072959 (PI James Leverenz, MD).}

\printcredits

\bibliographystyle{cas-model2-names}
\bibliography{main}

\begin{thebibliography}{49}
\expandafter\ifx\csname natexlab\endcsname\relax\def\natexlab#1{#1}\fi
\providecommand{\url}[1]{\texttt{#1}}
\providecommand{\href}[2]{#2}
\providecommand{\path}[1]{#1}
\providecommand{\DOIprefix}{doi:}
\providecommand{\ArXivprefix}{arXiv:}
\providecommand{\URLprefix}{URL: }
\providecommand{\Pubmedprefix}{pmid:}
\providecommand{\doi}[1]{\href{http://dx.doi.org/#1}{\path{#1}}}
\providecommand{\Pubmed}[1]{\href{pmid:#1}{\path{#1}}}
\providecommand{\bibinfo}[2]{#2}
\ifx\xfnm\relax \def\xfnm[#1]{\unskip,\space#1}\fi
\bibitem[{Beekly et~al.(2007)Beekly, Ramos, Lee, Deitrich, Jacka, Wu, Hubbard, Koepsell, Morris, Kukull et~al.}]{beekly2007national}
\bibinfo{author}{Beekly, D.L.}, \bibinfo{author}{Ramos, E.M.}, \bibinfo{author}{Lee, W.W.}, \bibinfo{author}{Deitrich, W.D.}, \bibinfo{author}{Jacka, M.E.}, \bibinfo{author}{Wu, J.}, \bibinfo{author}{Hubbard, J.L.}, \bibinfo{author}{Koepsell, T.D.}, \bibinfo{author}{Morris, J.C.}, \bibinfo{author}{Kukull, W.A.}, et~al., \bibinfo{year}{2007}.
\newblock \bibinfo{title}{The national alzheimer's coordinating center (nacc) database: the uniform data set}.
\newblock \bibinfo{journal}{Alzheimer Disease \& Associated Disorders} \bibinfo{volume}{21}, \bibinfo{pages}{249--258}.
\bibitem[{Billot et~al.(2023)Billot, Greve, Puonti, Thielscher, Van~Leemput, Fischl, Dalca, Iglesias et~al.}]{billot2023synthseg}
\bibinfo{author}{Billot, B.}, \bibinfo{author}{Greve, D.N.}, \bibinfo{author}{Puonti, O.}, \bibinfo{author}{Thielscher, A.}, \bibinfo{author}{Van~Leemput, K.}, \bibinfo{author}{Fischl, B.}, \bibinfo{author}{Dalca, A.V.}, \bibinfo{author}{Iglesias, J.E.}, et~al., \bibinfo{year}{2023}.
\newblock \bibinfo{title}{Synthseg: Segmentation of brain mri scans of any contrast and resolution without retraining}.
\newblock \bibinfo{journal}{Medical image analysis} \bibinfo{volume}{86}, \bibinfo{pages}{102789}.
\bibitem[{Birkenbihl et~al.(2024)Birkenbihl, de~Jong, Yalchyk and Fr{\"o}hlich}]{birkenbihl2024deep}
\bibinfo{author}{Birkenbihl, C.}, \bibinfo{author}{de~Jong, J.}, \bibinfo{author}{Yalchyk, I.}, \bibinfo{author}{Fr{\"o}hlich, H.}, \bibinfo{year}{2024}.
\newblock \bibinfo{title}{Deep learning-based patient stratification for prognostic enrichment of clinical dementia trials}.
\newblock \bibinfo{journal}{Brain Communications} \bibinfo{volume}{6}, \bibinfo{pages}{fcae445}.
\bibitem[{Blumberg et~al.(2018)Blumberg, Tanno, Kokkinos and Alexander}]{blumberg2018deeper}
\bibinfo{author}{Blumberg, S.B.}, \bibinfo{author}{Tanno, R.}, \bibinfo{author}{Kokkinos, I.}, \bibinfo{author}{Alexander, D.C.}, \bibinfo{year}{2018}.
\newblock \bibinfo{title}{Deeper image quality transfer: Training low-memory neural networks for 3d images}, in: \bibinfo{booktitle}{Medical Image Computing and Computer Assisted Intervention--MICCAI 2018: 21st International Conference, Granada, Spain, September 16-20, 2018, Proceedings, Part I}, \bibinfo{organization}{Springer}. pp. \bibinfo{pages}{118--125}.
\bibitem[{Cardoso et~al.(2022)Cardoso, Li, Brown, Ma, Kerfoot, Wang, Murrey, Myronenko, Zhao, Yang et~al.}]{cardoso2022monai}
\bibinfo{author}{Cardoso, M.J.}, \bibinfo{author}{Li, W.}, \bibinfo{author}{Brown, R.}, \bibinfo{author}{Ma, N.}, \bibinfo{author}{Kerfoot, E.}, \bibinfo{author}{Wang, Y.}, \bibinfo{author}{Murrey, B.}, \bibinfo{author}{Myronenko, A.}, \bibinfo{author}{Zhao, C.}, \bibinfo{author}{Yang, D.}, et~al., \bibinfo{year}{2022}.
\newblock \bibinfo{title}{Monai: An open-source framework for deep learning in healthcare}.
\newblock \bibinfo{journal}{arXiv preprint arXiv:2211.02701} .
\bibitem[{Chadebec et~al.(2022)Chadebec, Huijben, Pluim, Allassonni{\`e}re and van Eijnatten}]{chadebec2022image}
\bibinfo{author}{Chadebec, C.}, \bibinfo{author}{Huijben, E.M.}, \bibinfo{author}{Pluim, J.P.}, \bibinfo{author}{Allassonni{\`e}re, S.}, \bibinfo{author}{van Eijnatten, M.A.}, \bibinfo{year}{2022}.
\newblock \bibinfo{title}{An image feature mapping model for continuous longitudinal data completion and generation of synthetic patient trajectories}, in: \bibinfo{booktitle}{MICCAI Workshop on Deep Generative Models}, \bibinfo{organization}{Springer}. pp. \bibinfo{pages}{55--64}.
\bibitem[{Cury et~al.(2019)Cury, Durrleman, Cash, Lorenzi, Nicholas, Bocchetta, Van~Swieten, Borroni, Galimberti, Masellis et~al.}]{cury2019spatiotemporal}
\bibinfo{author}{Cury, C.}, \bibinfo{author}{Durrleman, S.}, \bibinfo{author}{Cash, D.M.}, \bibinfo{author}{Lorenzi, M.}, \bibinfo{author}{Nicholas, J.M.}, \bibinfo{author}{Bocchetta, M.}, \bibinfo{author}{Van~Swieten, J.C.}, \bibinfo{author}{Borroni, B.}, \bibinfo{author}{Galimberti, D.}, \bibinfo{author}{Masellis, M.}, et~al., \bibinfo{year}{2019}.
\newblock \bibinfo{title}{Spatiotemporal analysis for detection of pre-symptomatic shape changes in neurodegenerative diseases: Initial application to the genfi cohort}.
\newblock \bibinfo{journal}{NeuroImage} \bibinfo{volume}{188}, \bibinfo{pages}{282--290}.
\bibitem[{Ellis et~al.(2009)Ellis, Bush, Darby, De~Fazio, Foster, Hudson, Lautenschlager, Lenzo, Martins, Maruff et~al.}]{ellis2009australian}
\bibinfo{author}{Ellis, K.A.}, \bibinfo{author}{Bush, A.I.}, \bibinfo{author}{Darby, D.}, \bibinfo{author}{De~Fazio, D.}, \bibinfo{author}{Foster, J.}, \bibinfo{author}{Hudson, P.}, \bibinfo{author}{Lautenschlager, N.T.}, \bibinfo{author}{Lenzo, N.}, \bibinfo{author}{Martins, R.N.}, \bibinfo{author}{Maruff, P.}, et~al., \bibinfo{year}{2009}.
\newblock \bibinfo{title}{The australian imaging, biomarkers and lifestyle (aibl) study of aging: methodology and baseline characteristics of 1112 individuals recruited for a longitudinal study of alzheimer's disease}.
\newblock \bibinfo{journal}{International psychogeriatrics} \bibinfo{volume}{21}, \bibinfo{pages}{672--687}.
\bibitem[{Eshaghi et~al.(2021)Eshaghi, Young, Wijeratne, Prados, Arnold, Narayanan, Guttmann, Barkhof, Alexander, Thompson et~al.}]{eshaghi2021identifying}
\bibinfo{author}{Eshaghi, A.}, \bibinfo{author}{Young, A.L.}, \bibinfo{author}{Wijeratne, P.A.}, \bibinfo{author}{Prados, F.}, \bibinfo{author}{Arnold, D.L.}, \bibinfo{author}{Narayanan, S.}, \bibinfo{author}{Guttmann, C.R.}, \bibinfo{author}{Barkhof, F.}, \bibinfo{author}{Alexander, D.C.}, \bibinfo{author}{Thompson, A.J.}, et~al., \bibinfo{year}{2021}.
\newblock \bibinfo{title}{Identifying multiple sclerosis subtypes using unsupervised machine learning and mri data}.
\newblock \bibinfo{journal}{Nature communications} \bibinfo{volume}{12}, \bibinfo{pages}{2078}.
\bibitem[{Firth et~al.(2019)Firth, Primativo, Marinescu, Shakespeare, Suarez-Gonzalez, Lehmann, Carton, Ocal, Pavisic, Paterson et~al.}]{firth2019longitudinal}
\bibinfo{author}{Firth, N.C.}, \bibinfo{author}{Primativo, S.}, \bibinfo{author}{Marinescu, R.V.}, \bibinfo{author}{Shakespeare, T.J.}, \bibinfo{author}{Suarez-Gonzalez, A.}, \bibinfo{author}{Lehmann, M.}, \bibinfo{author}{Carton, A.}, \bibinfo{author}{Ocal, D.}, \bibinfo{author}{Pavisic, I.}, \bibinfo{author}{Paterson, R.W.}, et~al., \bibinfo{year}{2019}.
\newblock \bibinfo{title}{Longitudinal neuroanatomical and cognitive progression of posterior cortical atrophy}.
\newblock \bibinfo{journal}{Brain} \bibinfo{volume}{142}, \bibinfo{pages}{2082--2095}.
\bibitem[{Goodfellow et~al.(2020)Goodfellow, Pouget-Abadie, Mirza, Xu, Warde-Farley, Ozair, Courville and Bengio}]{goodfellow2020generative}
\bibinfo{author}{Goodfellow, I.}, \bibinfo{author}{Pouget-Abadie, J.}, \bibinfo{author}{Mirza, M.}, \bibinfo{author}{Xu, B.}, \bibinfo{author}{Warde-Farley, D.}, \bibinfo{author}{Ozair, S.}, \bibinfo{author}{Courville, A.}, \bibinfo{author}{Bengio, Y.}, \bibinfo{year}{2020}.
\newblock \bibinfo{title}{Generative adversarial networks}.
\newblock \bibinfo{journal}{Communications of the ACM} \bibinfo{volume}{63}, \bibinfo{pages}{139--144}.
\bibitem[{He et~al.(2024)He, Ang and Tward}]{he2024individualized}
\bibinfo{author}{He, R.}, \bibinfo{author}{Ang, G.}, \bibinfo{author}{Tward, D.}, \bibinfo{year}{2024}.
\newblock \bibinfo{title}{Individualized multi-horizon mri trajectory prediction for alzheimer's disease}.
\newblock \bibinfo{journal}{arXiv preprint arXiv:2408.02018} .
\bibitem[{Ho et~al.(2020)Ho, Jain and Abbeel}]{ho2020denoising}
\bibinfo{author}{Ho, J.}, \bibinfo{author}{Jain, A.}, \bibinfo{author}{Abbeel, P.}, \bibinfo{year}{2020}.
\newblock \bibinfo{title}{Denoising diffusion probabilistic models}.
\newblock \bibinfo{journal}{Advances in neural information processing systems} \bibinfo{volume}{33}, \bibinfo{pages}{6840--6851}.
\bibitem[{Hoopes et~al.(2022)Hoopes, Mora, Dalca, Fischl and Hoffmann}]{hoopes2022synthstrip}
\bibinfo{author}{Hoopes, A.}, \bibinfo{author}{Mora, J.S.}, \bibinfo{author}{Dalca, A.V.}, \bibinfo{author}{Fischl, B.}, \bibinfo{author}{Hoffmann, M.}, \bibinfo{year}{2022}.
\newblock \bibinfo{title}{Synthstrip: Skull-stripping for any brain image}.
\newblock \bibinfo{journal}{NeuroImage} \bibinfo{volume}{260}, \bibinfo{pages}{119474}.
\bibitem[{Jung et~al.(2021)Jung, Luna and Park}]{jung2021conditional}
\bibinfo{author}{Jung, E.}, \bibinfo{author}{Luna, M.}, \bibinfo{author}{Park, S.H.}, \bibinfo{year}{2021}.
\newblock \bibinfo{title}{Conditional gan with an attention-based generator and a 3d discriminator for 3d medical image generation}, in: \bibinfo{booktitle}{Medical Image Computing and Computer Assisted Intervention--MICCAI 2021: 24th International Conference, Strasbourg, France, September 27--October 1, 2021, Proceedings, Part VI 24}, \bibinfo{organization}{Springer}. pp. \bibinfo{pages}{318--328}.
\bibitem[{Kim and Ye(2022)}]{kim2022diffusion}
\bibinfo{author}{Kim, B.}, \bibinfo{author}{Ye, J.C.}, \bibinfo{year}{2022}.
\newblock \bibinfo{title}{Diffusion deformable model for 4d temporal medical image generation}, in: \bibinfo{booktitle}{International Conference on Medical Image Computing and Computer-Assisted Intervention}, \bibinfo{organization}{Springer}. pp. \bibinfo{pages}{539--548}.
\bibitem[{Kingma(2013)}]{kingma2013auto}
\bibinfo{author}{Kingma, D.P.}, \bibinfo{year}{2013}.
\newblock \bibinfo{title}{Auto-encoding variational bayes}.
\newblock \bibinfo{journal}{arXiv preprint arXiv:1312.6114} .
\bibitem[{Koval et~al.(2021)Koval, B{\^o}ne, Louis, Lartigue, Bottani, Marcoux, Samper-Gonzalez, Burgos, Charlier, Bertrand et~al.}]{koval2021ad}
\bibinfo{author}{Koval, I.}, \bibinfo{author}{B{\^o}ne, A.}, \bibinfo{author}{Louis, M.}, \bibinfo{author}{Lartigue, T.}, \bibinfo{author}{Bottani, S.}, \bibinfo{author}{Marcoux, A.}, \bibinfo{author}{Samper-Gonzalez, J.}, \bibinfo{author}{Burgos, N.}, \bibinfo{author}{Charlier, B.}, \bibinfo{author}{Bertrand, A.}, et~al., \bibinfo{year}{2021}.
\newblock \bibinfo{title}{{AD} course map charts alzheimer’s disease progression}.
\newblock \bibinfo{journal}{Scientific Reports} \bibinfo{volume}{11}, \bibinfo{pages}{8020}.
\bibitem[{LaMontagne et~al.(2019)LaMontagne, Benzinger, Morris, Keefe, Hornbeck, Xiong, Grant, Hassenstab, Moulder, Vlassenko et~al.}]{lamontagne2019oasis}
\bibinfo{author}{LaMontagne, P.J.}, \bibinfo{author}{Benzinger, T.L.}, \bibinfo{author}{Morris, J.C.}, \bibinfo{author}{Keefe, S.}, \bibinfo{author}{Hornbeck, R.}, \bibinfo{author}{Xiong, C.}, \bibinfo{author}{Grant, E.}, \bibinfo{author}{Hassenstab, J.}, \bibinfo{author}{Moulder, K.}, \bibinfo{author}{Vlassenko, A.G.}, et~al., \bibinfo{year}{2019}.
\newblock \bibinfo{title}{Oasis-3: longitudinal neuroimaging, clinical, and cognitive dataset for normal aging and alzheimer disease}.
\newblock \bibinfo{journal}{MedRxiv} , \bibinfo{pages}{2019--12}.
\bibitem[{Litrico et~al.(2024)Litrico, Guarnera, Giuffrida, Rav{\`\i} and Battiato}]{litrico2024tadm}
\bibinfo{author}{Litrico, M.}, \bibinfo{author}{Guarnera, F.}, \bibinfo{author}{Giuffrida, M.V.}, \bibinfo{author}{Rav{\`\i}, D.}, \bibinfo{author}{Battiato, S.}, \bibinfo{year}{2024}.
\newblock \bibinfo{title}{Tadm: Temporally-aware diffusion model for neurodegenerative progression on brain mri}, in: \bibinfo{booktitle}{International Conference on Medical Image Computing and Computer-Assisted Intervention}, \bibinfo{organization}{Springer}. pp. \bibinfo{pages}{444--453}.
\bibitem[{McMaster et~al.(2025)McMaster, Puglisi, Gao, Krishnan, Saunders, Ravi, Beason-Held, Resnick, Zuo, Moyer et~al.}]{mcmaster2025technical}
\bibinfo{author}{McMaster, E.}, \bibinfo{author}{Puglisi, L.}, \bibinfo{author}{Gao, C.}, \bibinfo{author}{Krishnan, A.R.}, \bibinfo{author}{Saunders, A.M.}, \bibinfo{author}{Ravi, D.}, \bibinfo{author}{Beason-Held, L.L.}, \bibinfo{author}{Resnick, S.M.}, \bibinfo{author}{Zuo, L.}, \bibinfo{author}{Moyer, D.}, et~al., \bibinfo{year}{2025}.
\newblock \bibinfo{title}{A technical assessment of latent diffusion for alzheimer's disease progression}, in: \bibinfo{booktitle}{Medical Imaging 2025: Image Processing}, \bibinfo{organization}{SPIE}. pp. \bibinfo{pages}{505--513}.
\bibitem[{Oxtoby and Alexander(2017)}]{oxtoby2017imaging}
\bibinfo{author}{Oxtoby, N.P.}, \bibinfo{author}{Alexander, D.C.}, \bibinfo{year}{2017}.
\newblock \bibinfo{title}{Imaging plus x: multimodal models of neurodegenerative disease}.
\newblock \bibinfo{journal}{Current opinion in neurology} \bibinfo{volume}{30}, \bibinfo{pages}{371}.
\bibitem[{Papamakarios et~al.(2021)Papamakarios, Nalisnick, Rezende, Mohamed and Lakshminarayanan}]{papamakarios2021normalizing}
\bibinfo{author}{Papamakarios, G.}, \bibinfo{author}{Nalisnick, E.}, \bibinfo{author}{Rezende, D.J.}, \bibinfo{author}{Mohamed, S.}, \bibinfo{author}{Lakshminarayanan, B.}, \bibinfo{year}{2021}.
\newblock \bibinfo{title}{Normalizing flows for probabilistic modeling and inference}.
\newblock \bibinfo{journal}{Journal of Machine Learning Research} \bibinfo{volume}{22}, \bibinfo{pages}{1--64}.
\bibitem[{Petersen et~al.(2010)Petersen, Aisen, Beckett, Donohue, Gamst, Harvey, Jack, Jagust, Shaw, Toga et~al.}]{petersen2010alzheimer}
\bibinfo{author}{Petersen, R.C.}, \bibinfo{author}{Aisen, P.S.}, \bibinfo{author}{Beckett, L.A.}, \bibinfo{author}{Donohue, M.C.}, \bibinfo{author}{Gamst, A.C.}, \bibinfo{author}{Harvey, D.J.}, \bibinfo{author}{Jack, C.R.}, \bibinfo{author}{Jagust, W.J.}, \bibinfo{author}{Shaw, L.M.}, \bibinfo{author}{Toga, A.W.}, et~al., \bibinfo{year}{2010}.
\newblock \bibinfo{title}{Alzheimer's disease neuroimaging initiative (adni): clinical characterization}.
\newblock \bibinfo{journal}{Neurology} \bibinfo{volume}{74}, \bibinfo{pages}{201--209}.
\bibitem[{Pinaya et~al.(2023)Pinaya, Graham, Kerfoot, Tudosiu, Dafflon, Fernandez, Sanchez, Wolleb, da~Costa, Patel et~al.}]{pinaya2023generative}
\bibinfo{author}{Pinaya, W.H.}, \bibinfo{author}{Graham, M.S.}, \bibinfo{author}{Kerfoot, E.}, \bibinfo{author}{Tudosiu, P.D.}, \bibinfo{author}{Dafflon, J.}, \bibinfo{author}{Fernandez, V.}, \bibinfo{author}{Sanchez, P.}, \bibinfo{author}{Wolleb, J.}, \bibinfo{author}{da~Costa, P.F.}, \bibinfo{author}{Patel, A.}, et~al., \bibinfo{year}{2023}.
\newblock \bibinfo{title}{Generative ai for medical imaging: extending the monai framework}.
\newblock \bibinfo{journal}{arXiv preprint arXiv:2307.15208} .
\bibitem[{Pinaya et~al.(2022)Pinaya, Tudosiu, Dafflon, Da~Costa, Fernandez, Nachev, Ourselin and Cardoso}]{pinaya2022brain}
\bibinfo{author}{Pinaya, W.H.}, \bibinfo{author}{Tudosiu, P.D.}, \bibinfo{author}{Dafflon, J.}, \bibinfo{author}{Da~Costa, P.F.}, \bibinfo{author}{Fernandez, V.}, \bibinfo{author}{Nachev, P.}, \bibinfo{author}{Ourselin, S.}, \bibinfo{author}{Cardoso, M.J.}, \bibinfo{year}{2022}.
\newblock \bibinfo{title}{Brain imaging generation with latent diffusion models}, in: \bibinfo{booktitle}{MICCAI Workshop on Deep Generative Models}, \bibinfo{organization}{Springer}. pp. \bibinfo{pages}{117--126}.
\bibitem[{Pini et~al.(2016)Pini, Pievani, Bocchetta, Altomare, Bosco, Cavedo, Galluzzi, Marizzoni and Frisoni}]{pini2016brain}
\bibinfo{author}{Pini, L.}, \bibinfo{author}{Pievani, M.}, \bibinfo{author}{Bocchetta, M.}, \bibinfo{author}{Altomare, D.}, \bibinfo{author}{Bosco, P.}, \bibinfo{author}{Cavedo, E.}, \bibinfo{author}{Galluzzi, S.}, \bibinfo{author}{Marizzoni, M.}, \bibinfo{author}{Frisoni, G.B.}, \bibinfo{year}{2016}.
\newblock \bibinfo{title}{Brain atrophy in alzheimer’s disease and aging}.
\newblock \bibinfo{journal}{Ageing research reviews} \bibinfo{volume}{30}, \bibinfo{pages}{25--48}.
\bibitem[{Pombo et~al.(2023)Pombo, Gray, Cardoso, Ourselin, Rees, Ashburner and Nachev}]{pombo2023equitable}
\bibinfo{author}{Pombo, G.}, \bibinfo{author}{Gray, R.}, \bibinfo{author}{Cardoso, M.J.}, \bibinfo{author}{Ourselin, S.}, \bibinfo{author}{Rees, G.}, \bibinfo{author}{Ashburner, J.}, \bibinfo{author}{Nachev, P.}, \bibinfo{year}{2023}.
\newblock \bibinfo{title}{Equitable modelling of brain imaging by counterfactual augmentation with morphologically constrained 3d deep generative models}.
\newblock \bibinfo{journal}{Medical Image Analysis} \bibinfo{volume}{84}, \bibinfo{pages}{102723}.
\bibitem[{Puglisi et~al.(2024)Puglisi, Alexander and Rav{\`\i}}]{puglisi2024enhancing}
\bibinfo{author}{Puglisi, L.}, \bibinfo{author}{Alexander, D.C.}, \bibinfo{author}{Rav{\`\i}, D.}, \bibinfo{year}{2024}.
\newblock \bibinfo{title}{Enhancing spatiotemporal disease progression models via latent diffusion and prior knowledge}, in: \bibinfo{booktitle}{International Conference on Medical Image Computing and Computer-Assisted Intervention}, \bibinfo{organization}{Springer}. pp. \bibinfo{pages}{173--183}.
\bibitem[{Ravi et~al.(2022)Ravi, Blumberg, Ingala, Barkhof, Alexander, Oxtoby, Initiative et~al.}]{ravi2022degenerative}
\bibinfo{author}{Ravi, D.}, \bibinfo{author}{Blumberg, S.B.}, \bibinfo{author}{Ingala, S.}, \bibinfo{author}{Barkhof, F.}, \bibinfo{author}{Alexander, D.C.}, \bibinfo{author}{Oxtoby, N.P.}, \bibinfo{author}{Initiative, A.D.N.}, et~al., \bibinfo{year}{2022}.
\newblock \bibinfo{title}{Degenerative adversarial neuroimage nets for brain scan simulations: Application in ageing and dementia}.
\newblock \bibinfo{journal}{Medical Image Analysis} \bibinfo{volume}{75}, \bibinfo{pages}{102257}.
\bibitem[{Rivera(2024)}]{rivera2024train}
\bibinfo{author}{Rivera, M.}, \bibinfo{year}{2024}.
\newblock \bibinfo{title}{How to train your vae}, in: \bibinfo{booktitle}{2024 IEEE International Conference on Image Processing (ICIP)}, \bibinfo{organization}{IEEE}. pp. \bibinfo{pages}{3882--3888}.
\bibitem[{Rombach et~al.(2022)Rombach, Blattmann, Lorenz, Esser and Ommer}]{rombach2022high}
\bibinfo{author}{Rombach, R.}, \bibinfo{author}{Blattmann, A.}, \bibinfo{author}{Lorenz, D.}, \bibinfo{author}{Esser, P.}, \bibinfo{author}{Ommer, B.}, \bibinfo{year}{2022}.
\newblock \bibinfo{title}{High-resolution image synthesis with latent diffusion models}, in: \bibinfo{booktitle}{Proceedings of the IEEE/CVF conference on computer vision and pattern recognition}, pp. \bibinfo{pages}{10684--10695}.
\bibitem[{Sauty and Durrleman(2022a)}]{sauty2022progression}
\bibinfo{author}{Sauty, B.}, \bibinfo{author}{Durrleman, S.}, \bibinfo{year}{2022}a.
\newblock \bibinfo{title}{Progression models for imaging data with longitudinal variational auto encoders}, in: \bibinfo{booktitle}{International Conference on Medical Image Computing and Computer-Assisted Intervention}, \bibinfo{organization}{Springer}. pp. \bibinfo{pages}{3--13}.
\bibitem[{Sauty and Durrleman(2022b)}]{sauty2022riemannian}
\bibinfo{author}{Sauty, B.}, \bibinfo{author}{Durrleman, S.}, \bibinfo{year}{2022}b.
\newblock \bibinfo{title}{Riemannian metric learning for progression modeling of longitudinal datasets}, in: \bibinfo{booktitle}{2022 IEEE 19th International Symposium on Biomedical Imaging (ISBI)}, \bibinfo{organization}{IEEE}. pp. \bibinfo{pages}{1--5}.
\bibitem[{Schiratti et~al.(2015)Schiratti, Allassonniere, Colliot and Durrleman}]{schiratti2015learning}
\bibinfo{author}{Schiratti, J.B.}, \bibinfo{author}{Allassonniere, S.}, \bibinfo{author}{Colliot, O.}, \bibinfo{author}{Durrleman, S.}, \bibinfo{year}{2015}.
\newblock \bibinfo{title}{Learning spatiotemporal trajectories from manifold-valued longitudinal data}.
\newblock \bibinfo{journal}{Advances in neural information processing systems} \bibinfo{volume}{28}.
\bibitem[{Schiratti et~al.(2017)Schiratti, Allassonni{\`e}re, Colliot and Durrleman}]{schiratti2017bayesian}
\bibinfo{author}{Schiratti, J.B.}, \bibinfo{author}{Allassonni{\`e}re, S.}, \bibinfo{author}{Colliot, O.}, \bibinfo{author}{Durrleman, S.}, \bibinfo{year}{2017}.
\newblock \bibinfo{title}{A bayesian mixed-effects model to learn trajectories of changes from repeated manifold-valued observations}.
\newblock \bibinfo{journal}{The Journal of Machine Learning Research} \bibinfo{volume}{18}, \bibinfo{pages}{4840--4872}.
\bibitem[{Shinohara et~al.(2014)Shinohara, Sweeney, Goldsmith, Shiee, Mateen, Calabresi, Jarso, Pham, Reich, Crainiceanu et~al.}]{shinohara2014statistical}
\bibinfo{author}{Shinohara, R.T.}, \bibinfo{author}{Sweeney, E.M.}, \bibinfo{author}{Goldsmith, J.}, \bibinfo{author}{Shiee, N.}, \bibinfo{author}{Mateen, F.J.}, \bibinfo{author}{Calabresi, P.A.}, \bibinfo{author}{Jarso, S.}, \bibinfo{author}{Pham, D.L.}, \bibinfo{author}{Reich, D.S.}, \bibinfo{author}{Crainiceanu, C.M.}, et~al., \bibinfo{year}{2014}.
\newblock \bibinfo{title}{Statistical normalization techniques for magnetic resonance imaging}.
\newblock \bibinfo{journal}{NeuroImage: Clinical} \bibinfo{volume}{6}, \bibinfo{pages}{9--19}.
\bibitem[{Song et~al.(2021)Song, Meng and Ermon}]{song2021denoising}
\bibinfo{author}{Song, J.}, \bibinfo{author}{Meng, C.}, \bibinfo{author}{Ermon, S.}, \bibinfo{year}{2021}.
\newblock \bibinfo{title}{Denoising diffusion implicit models}, in: \bibinfo{booktitle}{International Conference on Learning Representations}.
\bibitem[{Tijms et~al.(2024)Tijms, Vromen, Mjaavatten, Holstege, Reus, van~der Lee, Wesenhagen, Lorenzini, Vermunt, Venkatraghavan et~al.}]{tijms2024cerebrospinal}
\bibinfo{author}{Tijms, B.M.}, \bibinfo{author}{Vromen, E.M.}, \bibinfo{author}{Mjaavatten, O.}, \bibinfo{author}{Holstege, H.}, \bibinfo{author}{Reus, L.M.}, \bibinfo{author}{van~der Lee, S.}, \bibinfo{author}{Wesenhagen, K.E.}, \bibinfo{author}{Lorenzini, L.}, \bibinfo{author}{Vermunt, L.}, \bibinfo{author}{Venkatraghavan, V.}, et~al., \bibinfo{year}{2024}.
\newblock \bibinfo{title}{Cerebrospinal fluid proteomics in patients with alzheimer’s disease reveals five molecular subtypes with distinct genetic risk profiles}.
\newblock \bibinfo{journal}{Nature Aging} , \bibinfo{pages}{1--15}.
\bibitem[{Tustison et~al.(2010)Tustison, Avants, Cook, Zheng, Egan, Yushkevich and Gee}]{tustison2010n4itk}
\bibinfo{author}{Tustison, N.J.}, \bibinfo{author}{Avants, B.B.}, \bibinfo{author}{Cook, P.A.}, \bibinfo{author}{Zheng, Y.}, \bibinfo{author}{Egan, A.}, \bibinfo{author}{Yushkevich, P.A.}, \bibinfo{author}{Gee, J.C.}, \bibinfo{year}{2010}.
\newblock \bibinfo{title}{N4itk: improved n3 bias correction}.
\newblock \bibinfo{journal}{IEEE transactions on medical imaging} \bibinfo{volume}{29}, \bibinfo{pages}{1310--1320}.
\bibitem[{Vogel et~al.(2021)Vogel, Young, Oxtoby, Smith, Ossenkoppele, Strandberg, La~Joie, Aksman, Grothe, Iturria-Medina et~al.}]{vogel2021four}
\bibinfo{author}{Vogel, J.W.}, \bibinfo{author}{Young, A.L.}, \bibinfo{author}{Oxtoby, N.P.}, \bibinfo{author}{Smith, R.}, \bibinfo{author}{Ossenkoppele, R.}, \bibinfo{author}{Strandberg, O.T.}, \bibinfo{author}{La~Joie, R.}, \bibinfo{author}{Aksman, L.M.}, \bibinfo{author}{Grothe, M.J.}, \bibinfo{author}{Iturria-Medina, Y.}, et~al., \bibinfo{year}{2021}.
\newblock \bibinfo{title}{Four distinct trajectories of tau deposition identified in alzheimer’s disease}.
\newblock \bibinfo{journal}{Nature medicine} \bibinfo{volume}{27}, \bibinfo{pages}{871--881}.
\bibitem[{Wang et~al.(2024)Wang, Tachimori, Yamaguchi, Sekiguchi, Li, Yamashita and Initiative}]{wang2024multimodal}
\bibinfo{author}{Wang, C.}, \bibinfo{author}{Tachimori, H.}, \bibinfo{author}{Yamaguchi, H.}, \bibinfo{author}{Sekiguchi, A.}, \bibinfo{author}{Li, Y.}, \bibinfo{author}{Yamashita, Y.}, \bibinfo{author}{Initiative, A.D.N.}, \bibinfo{year}{2024}.
\newblock \bibinfo{title}{A multimodal deep learning approach for the prediction of cognitive decline and its effectiveness in clinical trials for alzheimer’s disease}.
\newblock \bibinfo{journal}{Translational psychiatry} \bibinfo{volume}{14}, \bibinfo{pages}{105}.
\bibitem[{Wilms et~al.(2022)Wilms, Bannister, Mouches, MacDonald, Rajashekar, Langner and Forkert}]{wilms2022invertible}
\bibinfo{author}{Wilms, M.}, \bibinfo{author}{Bannister, J.J.}, \bibinfo{author}{Mouches, P.}, \bibinfo{author}{MacDonald, M.E.}, \bibinfo{author}{Rajashekar, D.}, \bibinfo{author}{Langner, S.}, \bibinfo{author}{Forkert, N.D.}, \bibinfo{year}{2022}.
\newblock \bibinfo{title}{Invertible modeling of bidirectional relationships in neuroimaging with normalizing flows: application to brain aging}.
\newblock \bibinfo{journal}{IEEE Transactions on Medical Imaging} \bibinfo{volume}{41}, \bibinfo{pages}{2331--2347}.
\bibitem[{Xia et~al.(2021)Xia, Chartsias, Wang, Tsaftaris, Initiative et~al.}]{xia2021learning}
\bibinfo{author}{Xia, T.}, \bibinfo{author}{Chartsias, A.}, \bibinfo{author}{Wang, C.}, \bibinfo{author}{Tsaftaris, S.A.}, \bibinfo{author}{Initiative, A.D.N.}, et~al., \bibinfo{year}{2021}.
\newblock \bibinfo{title}{Learning to synthesise the ageing brain without longitudinal data}.
\newblock \bibinfo{journal}{Medical Image Analysis} \bibinfo{volume}{73}, \bibinfo{pages}{102169}.
\bibitem[{Yoon et~al.(2023)Yoon, Zhang, Suk, Guo and Li}]{yoon2023sadm}
\bibinfo{author}{Yoon, J.S.}, \bibinfo{author}{Zhang, C.}, \bibinfo{author}{Suk, H.I.}, \bibinfo{author}{Guo, J.}, \bibinfo{author}{Li, X.}, \bibinfo{year}{2023}.
\newblock \bibinfo{title}{Sadm: Sequence-aware diffusion model for longitudinal medical image generation}, in: \bibinfo{booktitle}{International Conference on Information Processing in Medical Imaging}, \bibinfo{organization}{Springer}. pp. \bibinfo{pages}{388--400}.
\bibitem[{Young et~al.(2018)Young, Marinescu, Oxtoby, Bocchetta, Yong, Firth, Cash, Thomas, Dick, Cardoso et~al.}]{young2018uncovering}
\bibinfo{author}{Young, A.L.}, \bibinfo{author}{Marinescu, R.V.}, \bibinfo{author}{Oxtoby, N.P.}, \bibinfo{author}{Bocchetta, M.}, \bibinfo{author}{Yong, K.}, \bibinfo{author}{Firth, N.C.}, \bibinfo{author}{Cash, D.M.}, \bibinfo{author}{Thomas, D.L.}, \bibinfo{author}{Dick, K.M.}, \bibinfo{author}{Cardoso, J.}, et~al., \bibinfo{year}{2018}.
\newblock \bibinfo{title}{Uncovering the heterogeneity and temporal complexity of neurodegenerative diseases with subtype and stage inference}.
\newblock \bibinfo{journal}{Nature communications} \bibinfo{volume}{9}, \bibinfo{pages}{4273}.
\bibitem[{Young et~al.(2024)Young, Oxtoby, Garbarino, Fox, Barkhof, Schott and Alexander}]{young2024data}
\bibinfo{author}{Young, A.L.}, \bibinfo{author}{Oxtoby, N.P.}, \bibinfo{author}{Garbarino, S.}, \bibinfo{author}{Fox, N.C.}, \bibinfo{author}{Barkhof, F.}, \bibinfo{author}{Schott, J.M.}, \bibinfo{author}{Alexander, D.C.}, \bibinfo{year}{2024}.
\newblock \bibinfo{title}{Data-driven modelling of neurodegenerative disease progression: thinking outside the black box}.
\newblock \bibinfo{journal}{Nature Reviews Neuroscience} , \bibinfo{pages}{1--20}.
\bibitem[{Zhang et~al.(2023)Zhang, Rao and Agrawala}]{zhang2023adding}
\bibinfo{author}{Zhang, L.}, \bibinfo{author}{Rao, A.}, \bibinfo{author}{Agrawala, M.}, \bibinfo{year}{2023}.
\newblock \bibinfo{title}{Adding conditional control to text-to-image diffusion models}, in: \bibinfo{booktitle}{Proceedings of the IEEE/CVF International Conference on Computer Vision}, pp. \bibinfo{pages}{3836--3847}.
\bibitem[{Zhou et~al.(2024)Zhou, Chen, Hou, Xie, Dvornek, Zhou, Wilson, Duncan, Liu and Zhou}]{zhou2024cascaded}
\bibinfo{author}{Zhou, Y.}, \bibinfo{author}{Chen, T.}, \bibinfo{author}{Hou, J.}, \bibinfo{author}{Xie, H.}, \bibinfo{author}{Dvornek, N.C.}, \bibinfo{author}{Zhou, S.K.}, \bibinfo{author}{Wilson, D.L.}, \bibinfo{author}{Duncan, J.S.}, \bibinfo{author}{Liu, C.}, \bibinfo{author}{Zhou, B.}, \bibinfo{year}{2024}.
\newblock \bibinfo{title}{Cascaded multi-path shortcut diffusion model for medical image translation}.
\newblock \bibinfo{journal}{Medical Image Analysis} \bibinfo{volume}{98}, \bibinfo{pages}{103300}.

\end{thebibliography}


\appendix

\section{Auxiliary models}\label{sec:app-aux-models}
In this section, we provide details about the different models used in the experiment section as auxiliary models.  

\subsection{Linear Model (LM)}\label{sec:app-aux-models-lm}
For the LM, we adopted a linear regression approach that minimizes the Huber loss (implemented as the HuberRegressor~\footnotemark[1] in the scikit-learn Python package) to reduce sensitivity to outliers. The model takes as input age $A$, age $B$, the subject's diagnosis, sex, and progression-related volumes $v_A$ at age $A$ to predict the progression-related volumes $v_B$ at age $B$.

\subsection{Disease Course Mapping (DCM)}\label{sec:app-aux-models-dcm}
We fit three logistic DCMs~\citep{schiratti2017bayesian} for each cognitive status considered in the study: CN, MCI, and AD. Each DCM estimates an average trajectory for the volumes of the regions included in the progression-related covariates. To predict the future progression-related covariates $c_B$ of a subject, the average trajectory is adjusted according to the observations $c_{A_1}, \dots, c_{A_n}$ from the $n$ past visits of the subject, generating an individual trajectory. Volumes are normalized to the range $[0,1]$. Since the DCM model requires increasing trends, volume series that exhibit decreasing trends are inverted (using $1-x$). The original implementation from the Leaspy~\footnotemark[2] Python package is used, and each model is optimized over 5000 iterations.

\footnotetext[1]{\url{https://scikit-learn.org/stable/modules/generated/sklearn.linear_model.HuberRegressor.html}}
\footnotetext[2]{\url{https://leaspy.readthedocs.io/en/stable/}}

\section{Dataset statements}\label{sec:dataset-statements}
\textit{(i) Data was collected by the AIBL study group. AIBL study methodology has been reported previously~\citep{ellis2009australian}
. (ii) Data used in the preparation of this article were obtained from the Alzheimer's Disease Neuroimaging Initiative (ADNI) database (\url{adni.loni.usc.edu}). The ADNI was launched in 2003 as a public-private partnership, led by Principal Investigator Michael W. Weiner, MD. The original goal of ADNI was to test whether serial MRI, positron emission tomography (PET), other biological markers, and clinical and neuropsychological assessment can be combined to measure the progression of MCI and early AD. The current goals include validating biomarkers for clinical trials, improving the generalizability of ADNI data by increasing diversity in the participant cohort, and to provide data concerning the diagnosis and progression of AD to the scientific community. For up-to-date information, see \url{adni.loni.usc.edu}.}

\begin{figure}[h!]
    \centering
    \includegraphics[width=0.5\linewidth]{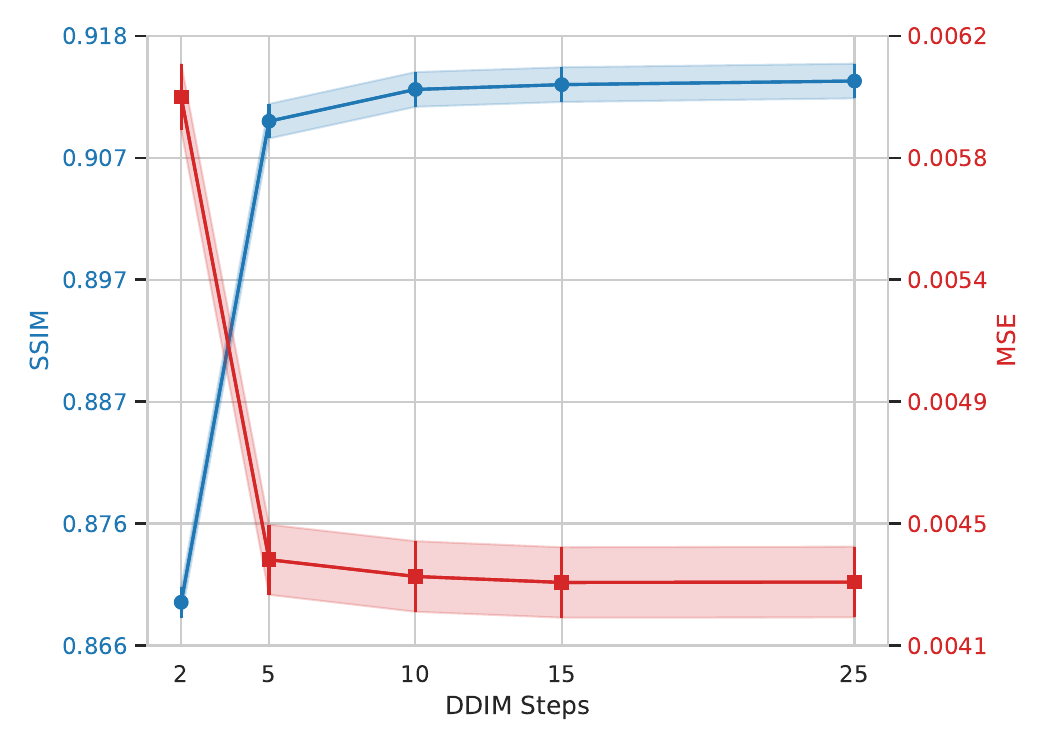}
    \caption{Effect of the number of DDIM inference steps on BrLP performance. SSIM (left axis, blue) and MSE (right axis, red) are reported for different numbers of denoising steps. Shaded areas indicate 95\% confidence intervals.}
    \label{fig:inference-steps}
\end{figure}

\section{Impact of reducing DDIM denoising steps on BrLP's performance}
\label{sec:ddim-steps}
We conduct an ablation study to evaluate how lowering the number of DDIM~\citep{song2021denoising} denoising steps affects BrLP's performance. Specifically, we compare the image-based metrics obtained using 25 steps (as in our experiments) to those obtained with 15, 10, 5, and 2 steps. The results, shown in Figure~\ref{fig:inference-steps}, reveal that reducing the number of steps to as few as 5 maintains satisfactory performance, with only a minor decline in quality.

\begin{table}[H]
    \caption{Evaluation of BrLP performance differences between male and female subjects. MSE and regional MAE values (± SD) are reported, with the best result for each metric between the two groups highlighted in bold. A result is marked with a star if it is significantly better than all other configurations at the 5\% significance level (Mann–Whitney U test).}
    \label{tab:sex-differences}
    \setlength{\tabcolsep}{5pt}
    \def\arraystretch{1.5}
    \resizebox{\columnwidth}{!}{
    
    \begin{tabular}{c|l|c|cc|ccc|cc} \hline 
    &&Exp. & \multicolumn{2}{c|}{\textbf{Image-based metrics}} & \multicolumn{3}{c|}{\textbf{MAE (conditional region volumes)}} & \multicolumn{2}{c}{\textbf{MAE (unconditional reg. volumes)}} \\
    &\textbf{Method}& Settings & MSE $\downarrow$ & SSIM $\uparrow$ & Hippocampus $\downarrow$ & Amygdala $\downarrow$ & Lat. Ventricle $\downarrow$ & Thalamus $\downarrow$ & CSF $\downarrow$ \\

    \hline
    \multirow{4}{*}{\rotatebox[origin=c]{90}{\parbox[c]{2cm}{\centering \textbf{\color{black}Internal\\test set\color{black}}}}}
    
    & Males & \multirow{2}{*}{\shortstack{Single\\ image}}  
    & 0.005 ± 0.003 & 0.909 ± 0.032 & \textbf{0.021 ± 0.019}* & 0.016 ± 0.014 & 0.273 ± 0.303 & 0.029 ± 0.023 & 0.828 ± 0.630 \\
    & Females & & \textbf{0.004 ± 0.002}* & \textbf{0.919 ± 0.022}* & 0.024 ± 0.022 & \textbf{0.014 ± 0.013} & \textbf{0.222 ± 0.294}* & \textbf{0.028 ± 0.022} & \textbf{0.823 ± 0.616} \\
    
    \cdashline{2-10}

    & Males & \multirow{2}{*}{\shortstack{Sequence\\ aware}} & 0.005 ± 0.003 & 0.910 ± 0.029 & 0.020 ± 0.018 & 0.015 ± 0.013 & 0.264 ± 0.279 & 0.031 ± 0.025 & 0.824 ± 0.658 \\
    & Females & & \textbf{0.004 ± 0.002}* & \textbf{0.918 ± 0.021}* & \textbf{0.020 ± 0.016} & \textbf{0.013 ± 0.012} & \textbf{0.199 ± 0.218}* & \textbf{0.029 ± 0.024} & \textbf{0.773 ± 0.577} \\

    \hline
    \multirow{4}{*}{\rotatebox[origin=c]{90}{\parbox[c]{2cm}{\centering \textbf{\color{black}External\\test set\color{black}}}}}
    
    & Males & \multirow{2}{*}{\shortstack{Single\\ image}} & 0.005 ± 0.002 & 0.906 ± 0.023 & 0.025 ± 0.021 & 0.015 ± 0.013 & \textbf{0.210 ± 0.240} & 0.030 ± 0.023 & \textbf{1.036 ± 0.736} \\
    & Females & & \textbf{0.004 ± 0.002}* & \textbf{0.910 ± 0.023}* & \textbf{0.024 ± 0.024} & \textbf{0.013 ± 0.013} & 0.215 ± 0.402 & \textbf{0.030 ± 0.024} & 1.049 ± 0.818 \\

    \cdashline{2-10}

    & Males & \multirow{2}{*}{\shortstack{Sequence\\ aware}} & 0.005 ± 0.002 & 0.910 ± 0.023 & \textbf{0.021 ± 0.018} & 0.014 ± 0.012 & 0.211 ± 0.234 & 0.030 ± 0.023 & 1.000 ± 0.718 \\
    & Females & & \textbf{0.004 ± 0.002}* & \textbf{0.913 ± 0.021}* & 0.022 ± 0.023 & \textbf{0.012 ± 0.012}* & \textbf{0.207 ± 0.430} & \textbf{0.030 ± 0.023} & \textbf{1.000 ± 0.775} \\

    \hline
    \end{tabular}
    }
\end{table}

\section{Analysis of sex differences in BrLP's predictions}
\label{sec:sex-differences}
To assess potential sex-related differences in BrLP’s performance, we stratify the evaluation by sex on both internal and external test sets, under single-image and sequence-aware settings. The results are summarized in Table~\ref{tab:sex-differences}. BrLP demonstrates slightly better performance for female subjects on image-based metrics in both test sets (Mann–Whitney U test, $p < 0.05$). Furthermore, volumetric analysis of the lateral ventricles shows a significantly lower MAE for female subjects in the internal test set ($p < 0.05$), although this difference does not persist in the external test set. In conclusion, we find that BrLP exhibits limited sex-related bias across both image-level and volumetric evaluations.

\end{document}